%% file: paper.tex
\documentclass[lettersize,journal]{IEEEtran}
\usepackage{times}

\usepackage[numbers]{natbib}
\usepackage{multicol}
\usepackage[bookmarks=true]{hyperref}
\usepackage{graphicx}

\usepackage{amsmath,amsfonts}
\usepackage{algorithm}
\usepackage{array}
\usepackage[caption=false,font=normalsize,labelfont=sf,textfont=sf]{subfig}
\usepackage{textcomp}
\usepackage{stfloats}
\usepackage{url}
\usepackage{verbatim}
\usepackage{amsmath,amssymb,amsopn,amstext,amsfonts}
\usepackage{color}
\usepackage{epstopdf}
\usepackage{setspace}
\usepackage{tikz}
\usepackage{multirow}
\usepackage{url}
\usepackage{verbatim}
\usepackage{siunitx} 
\usepackage{diagbox}
\usepackage{float}
\usepackage{epstopdf}
\usepackage{pifont}
\usepackage{fixltx2e}
\usepackage{balance}
\usepackage{color}
\usepackage{mathtools}
\usepackage{algpseudocode}
\usepackage{bm}

\usepackage{makecell}
\usepackage{enumerate}
\usepackage{booktabs}
\usepackage{algpseudocode}

\usepackage{xr}
\usepackage{cleveref}
\input{macros/macro.tex}

\input{macros/macro_alphabet.tex}
\begin{document}
	
	\title{WING: Wheel-Inertial Neural Odometry with Ground Manifold Constraints}
	
\author
{
	Chenxing Jiang$^{3,4,5}$, Kunyi Zhang$^{1,4,5,\dag}$, Sheng Yang$^{4}$, Shaojie Shen$^{3}$, Chao Xu$^{2,5}$, and Fei Gao$^{2,5,\dag}$
	\vspace{-0.3cm}\thanks{This work was supported by the National Key Research and Development Program of China (Grant NO. 2020AAA0108104), Alibaba Innovative Research (AIR) Program, and the National Natural Science Foundation of China under Grant 62003299.
		\big(\textit{Corresponding author: Kunyi Zhang, Fei Gao.}\big)}
	\vspace{-0.14cm}
	\thanks{\textsuperscript{1}The State Key Laboratory of Fluid Power and Mechatronic Systems, School of Mechanical Engineering, Zhejiang University, Hangzhou, China.}
	\thanks{\textsuperscript{2}State Key Laboratory of Industrial Control Technology, Institute of Cyber-Systems and Control, Zhejiang University, Hangzhou, China.}
	\thanks{\textsuperscript{3}Department of Electronic and Computer Engineering, Hong Kong University of Science and Technology.}
	\thanks{\textsuperscript{4}Alibaba DAMO Academy, Hangzhou, China.}
	\thanks{\textsuperscript{5}Huzhou Institute, Zhejiang University, Huzhou, China.}
	\thanks{E-mail: {cjiangan@connect.ust.hk, kunyizhang@zju.edu.cn}}
}

	\maketitle	
	\begin{abstract}
		In this paper, we propose an interoceptive-only odometry system for ground robots with neural network processing and soft constraints based on the assumption of a globally continuous ground manifold. Exteroceptive sensors such as cameras, GPS and LiDAR may encounter difficulties in scenarios with poor illumination, indoor environments, dusty areas and straight tunnels. Therefore, improving the pose estimation accuracy only using interoceptive sensors is important to enhance the reliability of navigation system even in degrading scenarios mentioned above.
		However, interoceptive sensors like IMU and wheel encoders suffer from large drift due to noisy measurements. To overcome these challenges, the proposed system trains deep neural networks to correct the measurements from IMU and wheel encoders, while considering their uncertainty. Moreover, because ground robots can only travel on the ground, we model the ground surface as a globally continuous manifold using a dual cubic B-spline manifold to further improve the estimation accuracy by this soft constraint. A novel space-based sliding-window filtering framework is proposed to fully exploit the $C^2$ continuity of ground manifold soft constraints and fuse all the information from raw measurements and neural networks in a yaw-independent attitude convention. Extensive experiments demonstrate that our proposed approach can outperform state-of-the-art learning-based interoceptive-only odometry methods.
	\end{abstract}
	\vspace{-0.5cm}
	\IEEEpeerreviewmaketitle
	\vspace{-0.95cm}
	\section{Introduction}
	\vspace{-0.1cm}
	Ground robots have been widely used in autonomous driving and space exploration. As a critical component of autonomous robot navigation, localization has been extensively studied for decades~\cite{chen2022milestones}. Most localization algorithms rely on exteroceptive sensors such as GPS~\cite{yusefi2023generalizable,wu2007gps}, LiDAR~\cite{li2020lidar, loam,lin2020loam}, and camera~\cite{geiger2012we, davison2007monoslam, klein2007parallel, mur2015orb, forster2014svo}. 
	Even though these methods can handle a wide range of conditions, exteroceptive sensors are still limited in certain degrading scenarios. For instance, GPS is unavailable in indoor scenes~\cite{yang2019observability} and cameras can be affected by poor illumination~\cite{wang2024four}. Besides, rain, fog and dust can impact LiDAR perception, while the repetitive feature pattern in straight tunnels can result in degraded pose estimation~\cite{wang2024four}. In contrast, interoceptive sensors like IMU or wheel encoders are not influenced by these environmental impacts. Therefore, improving the pose estimation accuracy only using interoceptive sensors is important to enhance the robustness and reliability of the overall navigation system in various real-world conditions.
	\begin{figure}[t!]
		\centering
		\includegraphics[width=0.95\linewidth]{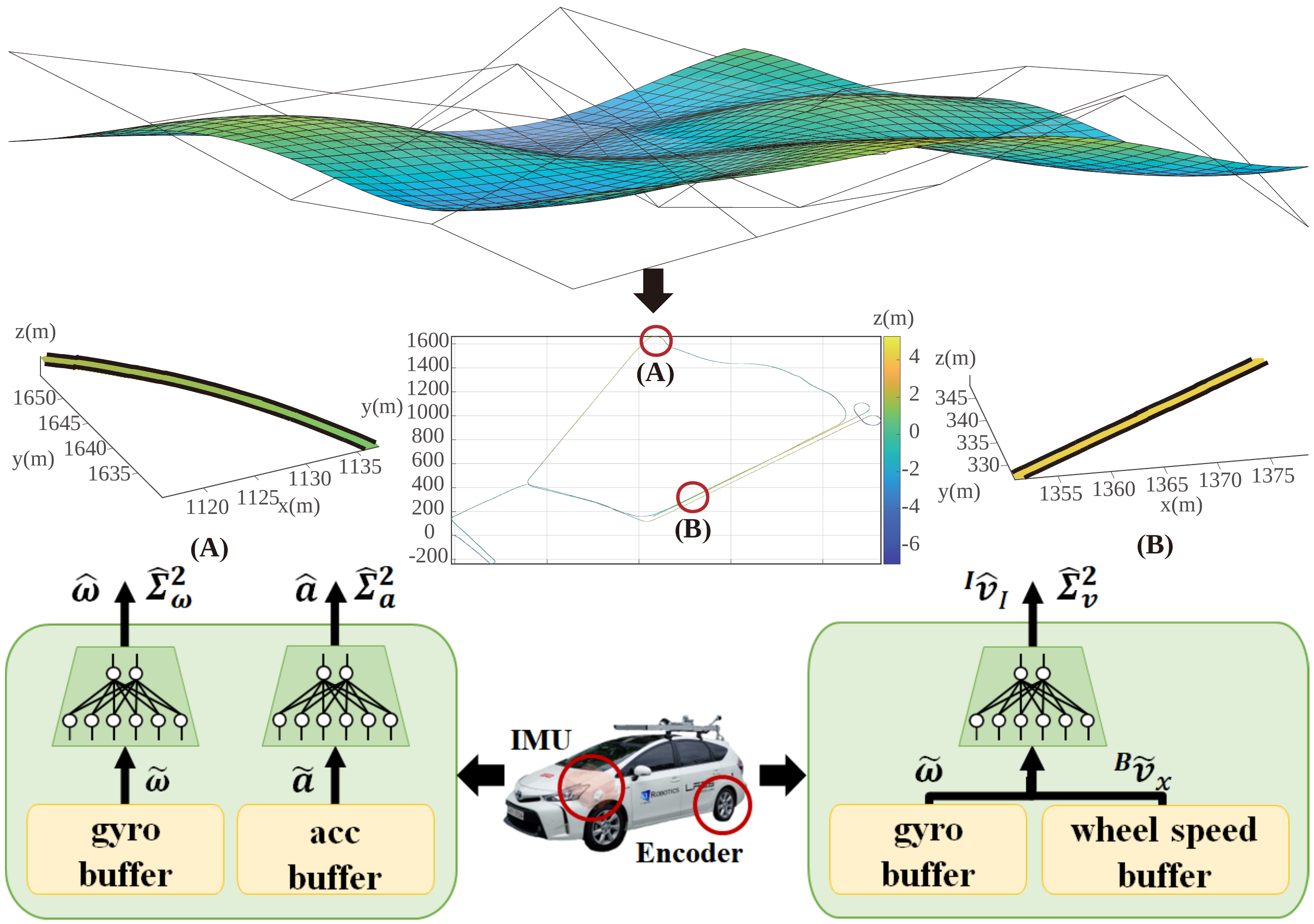}
		\caption{The pipeline of our proposed odometry. 
			Our proposed odometry estimates pose by processing the interoceptive sensor data through neural networks while simultaneously reconstructing a continuous cubic B-spline manifold. The middle row of the figure shows the estimated position and terrain of scenario \textit{urban17} in the KAIST Urban Dataset~\cite{jeong2019complex}, where \textbf{A} and \textbf{B} represent the corner and straight routes, respectively.}
		\label{fig:pipeline}
		\vspace{-0.5cm}
	\end{figure}
	
	However, relying solely on the raw measurements from interoceptive sensors like IMU and wheel encoders cannot provide accurate pose estimation for ground robots. IMU raw measurements often suffer from significant drift after several integrations without real-time calibration of intrinsic parameters~\cite{wisth2021unified}. On the other hand, extracting velocity from wheel encoders using a two-wheel~\cite{wu2017vins,zhang2019vision} or bicycle~\cite{kang2019vins,xiao2021lio} model relies on assumptions such as no skidding or vertical bouncing~\cite{brossard2019rins}, which may not hold in all scenarios. To deal with this issue, researchers have proposed data-driven approaches and demonstrate the potential to enhance the performance of inertial odometry~\cite{brossard2019rins,brossard2020ai,ronin2020Herath,tlio2020liu} or wheel-inertial odometry~\cite{brossard2019learning}. However, some methods~\cite{brossard2019rins, brossard2020ai} fail to account for errors originating from the IMU, limiting their effectiveness.
		Approaches predicting displacement~\cite{ronin2020Herath, tlio2020liu} or position errors~\cite{brossard2019learning} require neural networks to implicitly learn the kinematic model for state propagation. However, since the errors primarily stem from the sensor measurements rather than the kinematic model itself, this approach becomes redundant and can result in poorer representation and generalization capabilities. 
		Besides, fixed covariance in~\cite{brossard2019learning} restricts its capacity to effectively handle the uncertainty inherent in neural predictions. As a result, these methods fail to achieve desirable results, as shown in the experiment section (Sec.~\ref{exp:benchmark}).
		
		In addition to utilizing information from interoceptive sensors, recent works~\cite{zhang2019vision, zhang2021pose, wu2017vins, 9347674,10105944} have demonstrated that leveraging the ground manifold assumption can also enhance pose estimation for ground robots. Existing studies typically focus on segmented ground manifolds, which only provide locally discrete soft constraints. However, these segmented representations fail to fully exploit the information available from the entire ground manifold. Since neighboring patches of the ground manifold are often correlated with each other, a more comprehensive representation that captures the complete ground manifold can potentially improve the accuracy of pose estimation by leveraging these correlations.
		
		Therefore, in order to achieve an accurate interoceptive-only odometry, 
		we propose a system that fully utilizes the benefits from both data-driven methods and ground manifold soft constraints. In contrast to existing methods that integrate the kinematic model within neural networks~\cite{ronin2020Herath, tlio2020liu, brossard2019learning}, our approach retains the kinematic model and employ neural networks to only modify the error-prone raw measurements of IMU and wheel encoder. This enables us to implicitly perform the online calibration of these two sensors, and provide time-varying uncertainty prediction to achieve better sensor fusion. Additionally, to better portray the consistency and continuity of ground robots during their motion, we leverage a globally continuous cubic B-spline manifold to model the ground manifold to softly constrain the position and tilt of ground robots. 
		In contrast to previous works that utilize segmented planar manifolds~\cite{wu2017vins} or segmented curved manifolds~\cite{zhang2019vision}, the cubic B-spline manifold can provide globally continuous soft constraints rather than locally discrete soft constraints. Notably, the shared control points among neighboring patches~\cite{gordon1974b} facilitate the incorporation of historical spatial information and maintain consistency, allowing for better state estimation. Furthermore, we connect these two parts through a space-based sliding-window filtering framework that leverages the time-varying uncertainty prediction from neural networks. The proposed work contributes as follows:
	\begin{enumerate}
		\item The proposed system employs neural networks to correct the raw measurements from the IMU and wheel encoders. Besides, the time-varying uncertainty prediction improves sensor fusion;
		\item The proposed system parameterizes the ground manifold using a globally continuous cubic B-spline manifold. This manifold is used to enforce analytical soft constraints, which effectively capture the continuity of the robot's motion and correlations of neighboring patches of the ground manifold;
		\item The proposed system utilizes a space-based sliding-window filtering framework, which fuses all the information in an interoceptive-only odometry system to estimate the robot's poses along with the parameters of the globally continuous ground manifold.
	\end{enumerate}
	\vspace{-0.1cm}
	\section{Related works}
	\subsection{Odometry with wheel encoders or ground manifold aid}
	Over the past few decades, localization for ground robots has been heavily investigated. Some works fuse exteroceptive sensors with wheel encoders to get better performance. VIWO~\cite{lee2020visual} utilizes the wheel encoder based on the two-wheel vehicle model, where the robot's forward linear speed and angular speed are deduced from the two-wheel encoders. 
	Other works, such as VINS-vehicle~\cite{kang2019vins} and LIO-vehicle~\cite{xiao2021lio} use the bicycle model. These methods use the wheel encoder and steering angle sensor measurements to build a 2DOF vehicle dynamics model and then construct an additional pre-integration for the factor graph optimization. 
	
	Unlike aerial or underwater robots which can move freely in 3D space, ground robots can only travel on the ground. Therefore, some methods utilize the ground manifold as an additional soft constraint to improve the pose estimation accuracy~\cite{liang2002visual,scaramuzza20111, zhang2019vision,zhang2021pose,wu2017vins,9347674,10105944}. Based on the planar ground assumption, Liang~\cite{liang2002visual} solves the homography matrix to estimate the motion of ground robots. 
	Meanwhile, Scaramuzza~\cite{scaramuzza20111} proposes a one-point RANSAC outlier rejection method that speeds up pose estimation.  
	However, these approaches only focus on incorporating ground assumptions in visual data association. 
	Coupling the ground manifold assumption in the optimization problem, Wu~\cite{wu2017vins}, Chen~\cite{10105944} and Ouyang~\cite{9347674} add approximately planar manifold assumption to the system to reduce estimation error. 
	Recently, Zhang~\cite{zhang2019vision} parameterizes the ground in the form of a segmented quadratic manifold and fuse with the camera, IMU, and wheel encoder. 
	Furthermore, Zhang~\cite{zhang2021pose} estimates the 6D pose of robots and recovers the ground manifold only utilizing wheel encoders and camera, without IMU measurements. 
	However, these methods opt for a segmented manifold model that lacks the capability to incorporate historical information and ensure consistency between adjacent patches of ground manifold. 
	\vspace{-0.1cm}
	\subsection{Learning-based interoceptive-only odometry}
	With the tremendous improvement of deep learning, researchers nowadays begin to use data-driven approaches to improve the performance of interoceptive-only odometry.
	To enhance the accuracy of inertial odometry, LWOI~\cite{brossard2019learning}, AI-IMU~\cite{brossard2020ai}, RoNIN~\cite{ronin2020Herath}, and TLIO~\cite{tlio2020liu} utilize neural networks to obtain complete pose estimation only using IMU measurements. 
	Specifically, RINS-W~\cite{brossard2019rins} trains an LSTM network to detect specific motion pattern, which are then fused into an Iterative Extended Kalman Filtering (IEKF) framework to estimate robot states. 
	AI-IMU~\cite{brossard2020ai} uses a convolutional neural network (CNN) to dynamically output observation noise parameters of the invariant extended Kalman filtering framework. 
	RoNIN~\cite{ronin2020Herath} uses several different neural networks to regress a moving subject's horizontal positions and direction. 
	However, these methods~\cite{brossard2019rins, brossard2020ai} fail to account for errors originating from the IMU, limiting their effectiveness.
	TLIO~\cite{tlio2020liu} learns pseudo-measurements of relative displacement and places them as observations in a tightly coupled statistical cloning extended Kalman filtering. 
	To better fuse the measurements from wheel encoders,  
	LWOI~\cite{brossard2019learning} takes measurements from a Fiber Optical Gyroscope (FOG), wheel encoder, and IMU as inputs to train a Gaussian process that corrects dynamical and observation models, which are then fused in an Extended Kalman Filtering (EKF) framework. 
	However, these approaches~\cite{tlio2020liu, ronin2020Herath, brossard2019learning} predict displacement or position errors based on measurements from interoceptive sensors, which implies that these methods require neural networks to implicitly learn the kinematic model for state propagation. However, since the errors primarily stem from the sensor measurements rather than the kinematic model itself, this approach becomes redundant and can result in poorer representation and generalization capabilities. Besides, fixed covariance in~\cite{brossard2019learning} restricts its capacity to effectively handle the uncertainty inherent in neural predictions.
	\vspace{-0.1cm}
	\section{Preliminaries}
	\subsection{Wheel odometer model}
	In this paper, we consider a two-wheel ground robot model, which means that the forward velocity ${^\bodyframe}{v_x}$ and rotational angular velocity ${^\bodyframe}{{\omega}_z}$ of the robot can be derived from speeds of two wheels:
	\begin{equation}
		\begin{aligned}
			{^\bodyframe}{v_x} = \frac{\omega_l  r_l + \omega_r  r_r}{2}, \quad
			{^\bodyframe}{\omega_z} = \frac{\omega_r  r_r - \omega_l  r_l}{w_b},
		\end{aligned}
		\label{equ:wheel_odometer_model}
	\end{equation}
	where
	$\omega_l$ and $\omega_r$ are the speeds of the left and right wheels,
	$r_l$ and $r_r$ are the radii of two wheels,
	$w_b$ is the wheelbase between two wheels,
	${\bodyframe}$ is the body frame centered on the midpoint of two wheels (with a triaxial sequence of Front-Left-Up).
	\vspace{-0.1cm}
	\subsection{IMU kinematic model}
	IMU measurements include the non-gravitational acceleration $\widetilde{\accvel}$ and gyroscope $\widetilde{\rotvel}$, which are measured in the IMU frame $\imuframe$ (centered on the IMU sensor with a triaxial sequence of Front-Left-Up) and given by:
	\begin{equation}
		\begin{aligned}
			{^{\imuframe}}\widetilde{\accvel} &= 
			{_{\imuframe}^{\graframe}}\mathbf{R}^\top ({^{\graframe}}{\accvel} + {^{\graframe}}\boldsymbol{\mathrm{g}} ) +  
			{^{\imuframe}}{\bavel} + {\navel}, \\
			{^{\imuframe}}\widetilde{\rotvel} &= 
			{^{\imuframe}}{\rotvel} + {^{\imuframe}}{\bgvel} + {\ngvel},
		\end{aligned}
	\end{equation}
	where
	${^{\graframe}}{\accvel}$ is the true acceleration in the gravity-aligned frame ${\graframe}$~(with z-axis pointing up vertically),
	${^{\imuframe}}{\rotvel}$ is the true angular velocity in $\imuframe$ frame,
	${^{\graframe}}\boldsymbol{\mathrm{g}} = [0,0,-9.8 m/s^2]$ is the gravity vector in ${\graframe}$ frame,
	${_{\imuframe}^{\graframe}}\mathbf{R}$ is the rotation matrix from $\imuframe$ frame to ${\graframe}$ frame,
	${\navel}$ and $\nb_{\rotvel}$ are the additive Gaussian white noise in acceleration and gyroscope measurements,
	${\bavel}$ and ${\bgvel}$ are the biases of IMU modeled as random walks:
	\begin{equation}
		\begin{aligned}
			{\navel} &\sim \mathcal{N}(0,\Cov_{\accvel}^2), &
			\dot{\bias}_{\accvel} = {\nb_{{\bias}_{\accvel}}}
			&\sim \mathcal{N}(0,\Cov_{\bavel}^2), \\
			{\ngvel} &\sim \mathcal{N}(0,\Cov_{\rotvel}^2), &
			\dot{\bias}_{\rotvel} = {\nb_{{\bias}_{\rotvel}}}
			&\sim \mathcal{N}(0,\Cov_{\bgvel}^2),
		\end{aligned}
	\end{equation}
	where $\Cov_*^2$ is the covariance corresponding to each Gaussian distribution respectively.
	\vspace{-0.1cm}
	\subsection{Yaw independent attitude convention}
	\label{yaw_independent}
	Inspired by~\cite{svacha2019inertial}, we utilize an attitude convention, which decouples the attitude into yaw and tilt angles. 
	Therefore, the estimations of the velocity and the tilt angle could avoid being affected by the drift of the yaw angle when updating by body velocity. 
	To better utilize the wheel encoder for measuring body velocity, the attitude is represented as follows:
	\begin{equation}
		\label{att_mat}
		\begin{aligned}
			{}_{\imuframe}^{\graframe}\mathbf{R} &= {}_{\imuframe}^{\graframe}\mathbf{R}_\psi {}_{\imuframe}^{\graframe}\mathbf{R}_\phi, \\
			{}_{\imuframe}^{\graframe}\mathbf{R}_\psi &= {
				\left(
				\begin{array}{ccc}
					\cos\psi  & -\sin\psi & 0 \\ 
					\sin \psi & \cos \psi & 0 \\ 
					0 & 0 & 1 
				\end{array}
				\right)}, \\
			{}_{\imuframe}^{\graframe}\mathbf{R}_\phi 
			&= {
				\left(
				\begin{array}{ccc}
					\frac{1-s_{1}^2+s_{2}^2}{1+s_{1}^2+s_{2}^2} & \frac{-2 s_{1} s_{2}}{1+s_{1}^2+s_{2}^2} & \frac{2 s_{1}}{1+s_{1}^2+s_{2}^2} \\
					\frac{-2 s_{1} s_{2}}{1+s_{1}^2+s_{2}^2} & \frac{1+s_{1}^2-s_{2}^2}{1+s_{1}^2+s_{2}^2} & \frac{2 s_{2}}{1+s_{1}^2+s_{2}^2} \\ 
					\frac{-2 s_{1}}{1+s_{1}^2+s_{2}^2} & \frac{-2 s_{2}}{1+s_{1}^2+s_{2}^2} & \frac{1-s_{1}^2-s_{2}^2}{1+s_{1}^2+s_{2}^2}
				\end{array}
				\right)},
		\end{aligned}
	\end{equation}
	where ${}_{\imuframe}^{\graframe}\mathbf{R}$ is the rotation matrix from the IMU frame $\imuframe$ to the gravity-aligned frame $\graframe$. $\{s_1,s_2,\psi\}$ are the abbreviations of IMU rotation components $\{{}_{\imuframe}^{\graframe}s_1,{}_{\imuframe}^{\graframe}s_2,{}_{\imuframe}^{\graframe}\psi\}$. $\psi$ is the yaw angle, and $\mathbf{s} = (s_1, s_2)^\top$ are the stereographic coordinates corresponding to the tilt on the two-dimensional unit sphere $S^2$. 	
	\vspace{-0.1cm}
	\subsection{Uniform dual cubic B-spline manifold}
	Similar to the cubic B-spline curve, cubic B-spline manifolds exhibit global $C^2$ continuity, which means that they maintain the continuity and smoothness of position throughout the entire 3D spatial domain. Each cubic B-spline manifold is composed of cubic polynomial manifold patches, and each patch is affected locally by $16$ control points. These $16$ control points are arranged in a $4\times4$ matrix called control net~\cite{gordon1974b}. Notably, the shared control points among neighboring patches facilitate the incorporation of historical spatial information and maintain consistency, allowing for better state estimation. Referring to the content in~\cite{gordon1974b}, we consider the cubic B-spline manifold with the uniform knot vectors:
	\vspace{-0.1cm}
	\begin{equation}
		\begin{aligned}
			X &= \{x_1, x_2, \cdots, x_t~\vert~x_{i+1}-x_i 
			= \textrm{const}, 1 \leq i \leq t, i\in \mathbb{Z}\}, \\
			Y &= \{y_1, y_2, \cdots, y_s~\vert~y_{j+1}-y_j 
			= \textrm{const}, 1 \leq j \leq s, j\in \mathbb{Z}\}.
		\end{aligned}
	\end{equation}
	We represent the B-spline basis functions by the local parameter $u \in [0,1]$ and $v \in [0,1]$ instead of the global parameter $x \in [x_i,x_{i+1}]$ and $y \in [y_j,y_{j+1}]$ on each knot interval by performing the transformations as:
	\vspace{-0.1cm}
	\begin{equation}
		\begin{aligned}
			x &= x(u) = (1-u)x_i+ux_{i+1}, u \in [0,1],\\
			y &= y(v) = (1-v)y_j+vy_{j+1}, v \in [0,1].
		\end{aligned}
	\end{equation}
	\vspace{-0.1cm}
	Then, each patch of cubic B-spline manifold $\mathcal{S}(x,y)$
	can be represented in compact matrix form:
	\begin{equation}\label{local_representation}
		\begin{aligned}
			\mathcal{S}(x,y) &= \sum\nolimits_{i=0}^3 \sum\nolimits_{j=0}^3 B_{i,3}(x) B_{j,3}(y) C_{i,j}, \\
			\Rightarrow	\mathcal{S}(u,v) &= 
			\boldsymbol{u} \mathbf{B} \mathbf{C} \mathbf{B}^\top \boldsymbol{v}^\top,
		\end{aligned}
	\end{equation}
	\vspace{-0.1cm}
	where $\boldsymbol{u} = [u^3,  u^2,  u,  1]$, $\boldsymbol{v} = [v^3 , v^2 , v , 1]$, and
	\vspace{-0.1cm}
	\begin{equation}
		\label{surface_matrix}
		\begin{aligned}
			\mathbf{B} &= \frac{1}{6}
			\begin{bmatrix}
				-1 &  3 & -3 & 1 \\
				3 & -6 &  3 & 0 \\
				-3 &  0 &  3 & 0 \\
				1 &  4 &  1 & 0
			\end{bmatrix},
			\mathbf{C} =
			\begin{bmatrix}
				c_{0,0} & c_{0,1} & c_{0,2} & c_{0,3} \\
				c_{1,0} & c_{1,1} & c_{1,2} & c_{1,3} \\
				c_{2,0} & c_{2,1} & c_{2,2} & c_{2,3} \\
				c_{3,0} & c_{3,1} & c_{3,2} & c_{3,3}
			\end{bmatrix},
		\end{aligned}
	\end{equation}
	where $\mathbf{B}$ is the coefficient matrix of the cubic B-spline manifold, and $\mathbf{C}$ denotes the $4\times 4$ matrix called control net~\cite{gordon1974b}. Each element in $\mathbf{C}$ represents a control point that determines the corresponding patch of the cubic B-spline manifold.
	\vspace{-0.1cm}
	\section{Kinematic Neural Network Design}
	\subsection{\textit{IMU De-Bias Net}}
	\label{De-Bias Net}
	Accurately estimating IMU bias is crucial as multiple integrations from IMU measurements are required to obtain the velocity and position of the robot. 
	Zhang~\cite{zhang2022dido} demonstrates promising results in utilizing neural networks to estimate the bias of an IMU. However, an important aspect in a sensor fusion system is the inclusion of uncertainty, which is not addressed in~\cite{zhang2022dido}. In this study, we employ the~\textit{IMU De-Bias Net} to learn the kinematic characteristics of the IMU. This enables us to not only estimate the time-varying bias more accurately but also compute the corresponding covariance. By incorporating these uncertainty measures, we aim to enhance the performance of multi-sensor fusion.
	
	The~\textit{IMU De-Bias Net} comprises two components, one for the accelerometer and the other for the gyroscope. 
	Both components share the same network architecture, which is a 1D version of ResNet~\cite{he2016deep} with one residual block and two fully connected layers for output. Respectively, 
	input features are historical 
	raw accelerometer measurements ${^{\imuframe}}\widetilde{\accvel}$ and 
	raw gyroscope measurements ${^{\imuframe}} \widetilde{\rotvel}$ specific time interval.
	The outputs are 
	$\{{^{\imuframe}}\widehat{\bias}_{\accvel}, \boldsymbol{\zeta}_{{\boldsymbol v}_{i,i+n}}\}$ 
	and $\{{^{\imuframe}}\widehat{\bias}_{\rotvel}, \boldsymbol{\zeta}_{{\boldsymbol q}_{i,i+n}}\}$, which describe the bias and the covariance. Specifically, we assume that the covariance matrices $\widehat{\Cov}_{\boldsymbol{v}_{i,i+n}}^2$ and $\widehat{\Cov}_{\boldsymbol{q}_{i,i+n}}^2$ are respectively parameterized by the output vectors $\boldsymbol{\zeta}_{{\boldsymbol v}_{i,i+n}}$ and $\boldsymbol{\zeta}_{{\boldsymbol q}_{i,i+n}}$ using the following function:
		~$
		\widehat{\Cov}^2 = 
		\text{diag}(
		e^{2\widehat{\boldsymbol{\zeta}}_{x}}, 
		e^{2\widehat{\boldsymbol{\zeta}}_{y}},
		e^{2\widehat{\boldsymbol{\zeta}}_{z}}).
		$ 
	To optimize the \textit{IMU De-Bias Net}, the Negative Log Likelihood (NLL) loss functions of relative velocity and relative rotation (in quaternion) are defined on the following integrated increments:
	\begin{equation}
		\begin{aligned}
			\mathcal{L}_{\text{NLL},d\boldsymbol{v}} &= 
			\frac{1}{2N} \sum\nolimits_{i=1}^N
			\log\det(\widehat{\Cov}_{\boldsymbol{v}_{i,i+n}}^2) \\
			& +
			\frac{1}{2N}\sum\nolimits_{i=1}^N
			{\rVert 
				{\boldsymbol v}_{i,i+n} - 
				{\boldsymbol{\widehat v}}_{i,i+n}
				\rVert^2_{\widehat{\Cov}_{\boldsymbol{v}_{i,i+n}}^2}}, \\
			\mathcal{L}_{\text{NLL},d\boldsymbol{q}} &= 
			\frac{1}{2N}\sum\nolimits_{i=1}^N
			\log\det(\widehat{\Cov}_{\boldsymbol{q}_{i,i+n}}^2) \\
			& + 
			\frac{1}{2N}\sum\nolimits_{i=1}^N
			{\rVert 
				\mathrm{Log} \big(
				\left(
				{\boldsymbol{\widehat q}}_{i,i+n} \right)^* \otimes
				{\boldsymbol q}_{i,i+n} \big)
				\rVert^2_{\widehat{\Cov}_{\boldsymbol{q}_{i,i+n}}^2}},
			\label{nll_debias}
		\end{aligned}
	\end{equation}
	where
	\begin{subequations}
		\begin{align}
			{\boldsymbol{\widehat v}}_{i,i+n}
			&= \int_{i}^{i+n} 
			{_{\imuframe_t}^{\graframe}}\mathbf{R} 
			({^{\imuframe_t}}\widetilde{\accvel} - {^{\imuframe_t}}\boldsymbol{\mathrm{g}} - {^{\imuframe_t}}\widehat{\bias}_{\accvel}) dt, \\
			{\boldsymbol{v}}_{i,i+n}
			&= {^{\graframe}}\boldsymbol{v}_{\imuframe_{i+n}}-{^{\graframe}}\boldsymbol{v}_{\imuframe_i}, \\
			{\boldsymbol{\widehat q}}_{i,i+n}
			&= \int_{i}^{i+n}  
			\frac{1}{2} {_{\imuframe_t}^{\imuframe_{i}}} {\boldsymbol{q}} \otimes 
			({^{\imuframe_t}}\widetilde{\rotvel} - 
			{^{\imuframe_t}}\widehat{\bias}_{\rotvel}) dt,\\
			{{\boldsymbol q}}_{i,i+n}
			&= ({_{\imuframe_{i}}^{\graframe}} {\boldsymbol{q}})^* \otimes
			{_{\imuframe_{i+n}}^{\graframe}} {\boldsymbol{q}}.
		\end{align}
	\end{subequations}
	Note that, the subscript $i$ and $i+n$ are the abbreviation for time $t_i$ and $t_{i+n}$, 
	${^{\graframe}}\boldsymbol{v}_{\imuframe_{i}}$ is the velocity of IMU in frame $\graframe$,  
		${_{\imuframe_{i}}^{\graframe}} {\boldsymbol{q}}$ is the rotation from frame $\imuframe_{i}$ to frame $\graframe$,
	${\boldsymbol{\widehat v}}_{i,i+n}$ and 
	${\boldsymbol{\widehat q}}_{i,i+n}$ are the intermediate variables calculated from the estimated bias values
	${^{\imuframe}}\widehat{\bias}_{\accvel}$ and 
	${^{\imuframe}}\widehat{\bias}_{\rotvel}$ of the~\textit{IMU De-Bias Net} by the forward Euler method,
	${\boldsymbol{v}}_{i,i+n}$ and 
	${\boldsymbol{q}}_{i,i+n}$ are the ground truth velocity increment and relative rotation of each input sequence, $n$ is the window size,
	$\widehat{\Cov}_{\boldsymbol{v}_{i,i+n}}^2$ and
	$\widehat{\Cov}_{\boldsymbol{q}_{i,i+n}}^2$ are the estimated covariance matrices. 
	$N$ is the batch size,
	$*$ and $\otimes$ are the conjugate and multiplication of quaternion.
	In training, ${_{\imuframe}^{\graframe}}\mathbf{R}$ is the ground truth rotation. 
	The quaternion conjugate, quaternion multiplication, and quaternion logarithm $\mathrm{Log}$ operations are defined in~\cite{sola2017quaternion}. 
	\vspace{-0.1cm}
	\subsection{\textit{Wheel Encoder Net}}
	\label{Wheel Odometer Net}
	Commonly, it is assumed that there is no speed in the non-forward direction:
	\begin{equation}
		^{\bodyframe}\boldsymbol{\widetilde{v}}_{\bodyframe} = 
		\begin{bmatrix}
			^{\bodyframe}\widetilde{v}_{\bodyframe_x} & 0 & 0
		\end{bmatrix}^\top,
	\end{equation}
	where $^{\bodyframe}\widetilde{v}_{\bodyframe_x}$ is the measurement value from wheel encoder. 
	However, the above nonholonomic assumptions are not always strictly satisfied due to skidding and vertical bouncing.
	In addition, the extrinsic parameters
	$\{{{}_{\bodyframe}^{\imuframe}}\mathbf{R}, {}^{\imuframe}\boldsymbol{t}_{\bodyframe}\}$
	between the IMU frame ($\imuframe$) and the wheel frame ($\bodyframe$) require real-time calibration due to the effect of the vibrations and suspension system in motion.
	We aim to incorporate both IMU data and wheel encoder measurements into a neural network to handle wheel speed outliers and time-varying extrinsic parameters. The architecture of the \textit{Wheel Encoder Net} is the same as that of the \textit{IMU De-bias Net}.
	Considering the following velocity transfer equation:
	\begin{equation}
		\begin{aligned}
			{}^{\imuframe}{\boldsymbol{v}}_{\imuframe} &=
			{{}_{\bodyframe}^{\imuframe}}\mathbf{R} ~
			{}^{\bodyframe}{\boldsymbol{v}}_{\bodyframe} - 
			{}^{\imuframe}{\rotvel} \times {}^{\imuframe}\boldsymbol{t}_{\bodyframe}-
			{}^{\imuframe}\dot{\boldsymbol{t}}_{\bodyframe},
		\end{aligned}
	\end{equation}
	We feed historical raw wheel speed $^{\bodyframe}\widetilde{v}_{\bodyframe_x}$ and
	raw gyroscope measurements ${^{\imuframe}} \widetilde{\rotvel}$ over specific time interval into the \textit{Wheel Encoder Net} and output the compensation of the $^{\bodyframe}\boldsymbol{\widetilde{v}}_{\bodyframe}$ at the current time.
	To train the \textit{Wheel Encoder Net}, we define the negative log-likelihood (NLL) loss function of the body velocity as follows:
	\begin{equation}
		\begin{aligned}
			\mathcal{L}_{\text{NLL},\boldsymbol{v}} &= 
			\frac{1}{2N}\sum\nolimits_{i=1}^N
			\log\det(\widehat{\Cov}_{\boldsymbol{v}_{i}}^2) \\ &+
			\frac{1}{2N}\sum\nolimits_{i=1}^N
			{\rVert 
				{^{\imuframe}}{\boldsymbol{v}}{_{\imuframe}}_i-
				{^{\imuframe}}{\boldsymbol{\widehat{v}}}{_{\imuframe}}_i
				\rVert^2_{\widehat{\Cov}_{\boldsymbol{v}_{i}}^2}}, 
		\end{aligned}
	\end{equation}
	where
	\begin{equation}
		\label{velocity_corrected}
		\begin{aligned}
			{^{\imuframe}}{\boldsymbol{\widehat{v}}}{_{\imuframe}}_i &=  {{}_{\bodyframe}^{\imuframe}}\mathbf{R}~
			{}^{\bodyframe}\boldsymbol{\widetilde{v}}{_{\bodyframe}}_i +
			{}^{\imuframe}\delta\boldsymbol{\widehat{v}}{_{\imuframe}}_i, \\
			\widehat{\Cov}_{\boldsymbol{v}_{i}}^2 &= 
			\text{diag}(
			e^{{2\widehat{\boldsymbol{\zeta}}_{v_{i,x}}}}, 
			e^{2\widehat{\boldsymbol{\zeta}}_{v_{i,y}}},
			e^{2\widehat{\boldsymbol{\zeta}}_{v_{i,z}}})
		\end{aligned}
	\end{equation}
	and
	$\{{}^{\imuframe}\delta\boldsymbol{\widehat{v}}{_{\imuframe}}_i, \widehat{\boldsymbol{\zeta}}_{v_{i}}\}$ are the outputs of \textit{Wheel Encoder Net},
	${^{\imuframe}}{\boldsymbol{v}}{_{\imuframe}}$ is the ground truth velocity in the $\imuframe$ frame. 
	\section{Ground Manifold soft Constraint}
	\label{Ground_manifold_constraint}
	\subsection{Dual cubic B-spline manifold}
	Assuming the ground can be represented as a dual cubic B-spline manifold, any three-dimensional (3D) point $\boldsymbol{p}=(x,y,z)$ on the manifold satisfies the following equation:
	\begin{equation}
		\begin{aligned}
			z = \sum\nolimits_{i=0}^3 \sum\nolimits_{j=0}^3 B_{i,3}(x) B_{j,3}(y) c_{i,j}.
			\label{z=f(x,y)}
		\end{aligned}
	\end{equation}
	Denoting the manifold as $\mathcal{M}$, we can express the soft constraint equation as:
	\begin{equation}
		\label{manifold_xy}
		\begin{aligned}
			\mathcal{M}(\boldsymbol{p}) = 
			\sum\nolimits_{i=0}^3 \sum\nolimits_{j=0}^3 B_{i,3}(x) B_{j,3}(y) c_{i,j} - z = 0.
		\end{aligned}
	\end{equation}
	By using the local representation in Eq.~\eqref{local_representation}, we can express Eq.~\eqref{manifold_xy} in terms of the local coordinates $(u,v)$, where each patch of the cubic B-spline manifold is represented as follows:
	\begin{equation}
		\begin{aligned}
			\mathcal {M}(\boldsymbol{p}) 
			&= \sum\nolimits_{i=0}^3 \sum\nolimits_{j=0}^3 B_{i,3}(u) B_{j,3}(v) c_{i,j} -z = 0\\
			&= \boldsymbol{u} \mathbf{B} \mathbf{C} \mathbf{B}^\top \boldsymbol{v}^\top-z = 0,
		\end{aligned}
		\label{manifold_uv}
	\end{equation}	
	where 
	\begin{equation}
		\begin{aligned}
			u &= \frac{x-x_i}{x_{i+1}-x_i}, x\in [x_i,x_{i+1}],  \\
			v &= \frac{y-y_j}{y_{j+1}-y_j}, y\in [y_j,y_{j+1}].
		\end{aligned}
	\end{equation}
	Assuming that the uniform knot vector of the dual cubic B-spline manifold has a constant interval of $d = {x_{i+1}-x_i} = {y_{j+1}-y_j}$, we can express the transformation as follows:
	\begin{equation}
		\begin{aligned}
			u &= k_x x+b_x \in [0,1], ~ k_x = 1/{d}, ~ b_x = -g_x/{d},\\
			v &= k_y y+b_y \in [0,1], ~ k_y = 1/{d}, ~ b_y = -g_y/{d},
		\end{aligned}
	\end{equation}
	where $(g_x,g_y) = (x_i,y_j)$ is the global coordinate of the ground manifold.
	Therefore, Eq.~\eqref{manifold_uv} can be abbreviated as follows:
	\begin{equation}
		\begin{aligned}
			\mathcal {M}(\boldsymbol{p}) 
			&= \boldsymbol{u} \mathbf{B} \mathbf{C} \mathbf{B}^\top \boldsymbol{v}^\top-z = 0, \\
			&= \boldsymbol{x} \mathbf{K}_x \mathbf{B} \mathbf{C} \mathbf{B}^\top \mathbf{K}_y^\top \boldsymbol{y}^\top-z = 0,
		\end{aligned}
		\label{manifold_simple}
	\end{equation}
	where
	\begin{equation}
		\label{K_xK_y}
		\begin{aligned}
			\boldsymbol{x} =
			\begin{bmatrix}
				x^3 , x^2 , x , 1
			\end{bmatrix},
			\mathbf{K}_x =        
			\begin{bmatrix}
				k_x^3     & 0 		& 0   & 0 \\
				3k_x^2b_x & k_x^2 	& 0   & 0 \\
				3k_xb_x^2 & 2k_xb_x & k_x & 0 \\
				b_x^3 	  & b_x^2 	& b_x & 1
			\end{bmatrix},\\
			\boldsymbol{y} =
			\begin{bmatrix}
				y^3 , y^2 , y , 1
			\end{bmatrix},
			\mathbf{K}_y =        
			\begin{bmatrix}
				k_y^3     & 0 		& 0   & 0 \\
				3k_y^2b_y & k_y^2 	& 0   & 0 \\
				3k_yb_y^2 & 2k_yb_y & k_y & 0 \\
				b_y^3 	  & b_y^2 	& b_y & 1
			\end{bmatrix}.
		\end{aligned}
	\end{equation}
	\subsection{Ground manifold soft constraint}
	Furthermore, given that the robot travels on the ground that is parameterized as a dual cubic B-spline manifold as Eq.~\eqref{manifold_simple}, which indicates that the poses of the point of contact between the wheels and the ground satisfy two kinds of soft constraints, one is that the position where the wheel attached to the ground ${}^{\graframe}\boldsymbol{p}_w $ is on the ground manifold, and the other is that the orientation ${}_{w}^{\graframe}\mathbf{R}$ is parallel to the normal of the ground manifold:
	\begin{equation}
		\label{manifold_constraint}
		\begin{aligned}
			&\mathcal{M} \left({}^{\graframe}\boldsymbol{p}_w \right) = 0, \\
			&{}_{w}^{\graframe}\mathbf{R} \cdot {\boldsymbol{e}_3} \times
			\nabla \mathcal{M} \left({}^{\graframe}\boldsymbol{p}_w \right) = \mathbf{0},
		\end{aligned}
	\end{equation}
	where ${\boldsymbol{e}_3} = [0,0,1]{^\top}$ is the third axis,
	$\nabla = \frac{\partial}{\partial x}\vec{i} + \frac{\partial}{\partial y}\vec{j} + \frac{\partial}{\partial z}\vec{k}$ is a vector differential operator,
	and $\mathbf{0}$ is the zero vector.
	The pose of any point of contact between the wheel and the ground
	$\{{}_{w}^{\graframe}\mathbf{R},{}^{\graframe}\boldsymbol{p}_{w}\}$
	can be converted by extrinsic transformation 
	$\{{}_{\imuframe}^{w}\mathbf{R},{}^{\imuframe}\boldsymbol{t}_{w}\}$
	from the IMU pose $\{{}_{\imuframe}^{\graframe}\mathbf{R},{}^{\graframe}\boldsymbol{p}_{\imuframe}\}$:
	\begin{equation}
		\label{wheel_pose}
		\begin{aligned}
			{}^{\graframe}\boldsymbol{p}_w &= {}^{\graframe}\boldsymbol{p}_{\imuframe} + {}_{\imuframe}^{\graframe}\mathbf{R}~ {}^{\imuframe}\boldsymbol{t}_{w}, \\
			{}_{w}^{\graframe}\mathbf{R} &= {}_{\imuframe}^{\graframe}\mathbf{R}~{}_{w}^{\imuframe}\mathbf{R}.
		\end{aligned}
	\end{equation}
	Assuming that the $\imuframe$ frame and the ${w}$ frame (centered on the point of contact between the wheel and the ground, with a triaxial sequence of Front-Left-Up) have the same rotation representation, 
	we can substitute Eq.~\eqref{att_mat}, Eq.~\eqref{manifold_simple} and Eq.~\eqref{wheel_pose} into equation Eq.~\eqref{manifold_constraint}.
	After simplification, we obtain:
	\begin{equation}
		\label{manifold_constraint_simple}
		\begin{aligned}
			\boldsymbol{x}_w \mathbf{K}_x \mathbf{B} \mathbf{C} \mathbf{B}^\top \mathbf{K}_y^\top \boldsymbol{y}^\top &- z_w = 0, \\ 
			\boldsymbol{x}_w \mathbf{K}_x \mathbf{B} \mathbf{C} \mathbf{B}^\top \mathbf{K}_y^\top \boldsymbol{\partial y}_w^\top &+ 2 \frac{s_1 \sin(\psi) + s_2 \cos(\psi)}{1-s_1^2-s_2^2} = 0,\\ 
			\boldsymbol{\partial x}_w \mathbf{K}_x \mathbf{B} \mathbf{C} \mathbf{B}^\top \mathbf{K}_y^\top \boldsymbol{y}_w^\top &+ 2 \frac{s_1 \cos(\psi) - s_2 \sin(\psi)}{1-s_1^2-s_2^2} = 0,
		\end{aligned}
	\end{equation}
	where 
	\begin{equation}
		\begin{aligned}
			\boldsymbol{x}_w &= 
			\begin{bmatrix}
				x_w^3, x_w^2, x_w, 1
			\end{bmatrix},
			\boldsymbol{\partial x}_w = 
			\begin{bmatrix}
				3x_w^2, 2x_w, 1, 0
			\end{bmatrix},\\
			\boldsymbol{y_w} &= 
			\begin{bmatrix}
				y_w^3, y_w^2, y_w, 1
			\end{bmatrix},~
			\boldsymbol{\partial y_w} = 
			\begin{bmatrix}
				3y_w^2, 2y_w, 1, 0
			\end{bmatrix},
		\end{aligned}
	\end{equation}
	The details of formula derivation can be found in Appendix~\ref{derivation_obs}. ${}^{\graframe}\boldsymbol{p}_w = [x_w, y_w, z_w]^\top$ is calculated by Eq.~\eqref{wheel_pose},
	$\mathbf{K}_x$, $\mathbf{K}_y$, $\mathbf{B}$ and $\mathbf{C}$ have the same definition as Eq.~\eqref{K_xK_y}, 
	$\{s_1,s_2,\psi\}$ are the abbreviations of IMU rotation components $\{{}_{\imuframe}^{\graframe}s_1,{}_{\imuframe}^{\graframe}s_2,{}_{\imuframe}^{\graframe}\psi\}$.
	Furthermore, we can transfer Eq.~\eqref{manifold_constraint_simple} to a vector form w.r.t. the matrix $\mathbf{C}$:
	\begin{equation}
		\label{manifold_constraint_concise}
		\resizebox{0.95\hsize}{!}{$
			\begin{aligned}
				(\boldsymbol{y}_w \mathbf{K}_y \mathbf{B} \boxtimes
				\boldsymbol{x}_w \mathbf{K}_x \mathbf{B})&Vec(\mathbf{C}) - z = 0,\\
				(\boldsymbol{\partial y^\top}_w \mathbf{K}_y \mathbf{B} \boxtimes
				\boldsymbol{x}_w \mathbf{K}_x \mathbf{B})&Vec(\mathbf{C}) + 2 \frac{s_1 \sin(\psi) + s_2 \cos(\psi)}{1-s_1^2-s_2^2} = 0,\\ 
				(\boldsymbol{y}_w^\top \mathbf{K}_y \mathbf{B} \boxtimes
				\boldsymbol{\partial x}_w \mathbf{K}_x \mathbf{B})&Vec(\mathbf{C}) + 2 \frac{s_1 \cos(\psi) - s_2 \sin(\psi)}{1-s_1^2-s_2^2} = 0,\\
				Vec(\mathbf{C}) = [c_{0,0}, \cdots , &c_{3,0}, \cdots , c_{0,3}, \cdots, c_{3,3}], \quad \mathbf{C} \in \mathbb{R}^{4\times4}
			\end{aligned}$}
	\end{equation}
	where $\boxtimes$ is the Kronecker product, and $Vec(\mathbf{C})$ is the column straightening function that transforms a matrix into a vector.
	
	\section{Space-based sliding-window fusion}
	To better estimate the poses of a ground robot, we apply a space-based sliding-window framework that combines manifold soft constraints with neural-processed IMU and wheel odometer data. This approach is specifically designed to not only smooth poses within a sliding window but also recover the parameters of the dual cubic B-spline manifold.
	\subsection{States}
	The state vector $\mathcal{X}$ includes IMU state $\mathcal{X}_I$, space-based sliding-window states $\mathcal{X}_S$ and space-based sliding control vector $\boldsymbol{c}$, which is represented as follows:
	\begin{equation}
		\begin{aligned}
			\mathcal{X} &= 
			\begin{bmatrix}
				\mathcal{X}_I {}^\top &
				\mathcal{X}_S {}^\top &
				\boldsymbol{c} {}^\top
			\end{bmatrix} {}^\top, \label{state_vec}\\
			\mathcal{X}_I &=
			\begin{bmatrix}
				^{\graframe}\boldsymbol{p}_{\imuframe} {}^\top & 
				^{\imuframe}\boldsymbol{v}_{\imuframe} {}^\top & 
				^{\graframe}_{\imuframe}{\psi} &
				^{\graframe}_{\imuframe}\boldsymbol{s} {}^\top 
			\end{bmatrix} {}^\top,  \\
			\mathcal{X}_S &=
			\begin{bmatrix}
				\boldsymbol{\xi}_1 &
				\boldsymbol{\xi}_2 &
				\cdots &
				\boldsymbol{\xi}_n
			\end{bmatrix} {}^\top, \\
			\boldsymbol{\xi} &=
			\begin{bmatrix}
				^{\graframe}\boldsymbol{p}_{\imuframe}^\top & 
				^{\graframe}_{\imuframe}{\psi} &
				^{\graframe}_{\imuframe}\boldsymbol{s}^\top
			\end{bmatrix} {}^\top,\\
			\boldsymbol{c} &= Vec(\mathbf{C}),  
		\end{aligned}
	\end{equation}
	where
	$^{\graframe}\boldsymbol{p}_{\imuframe}$,
	$^{\graframe}_{\imuframe}{\psi}$ and
	$^{\graframe}_{\imuframe}\boldsymbol{s}$ are the position, yaw angle, and tilt vector of the ground robot in the ${\graframe}$ frame,
	$^{\imuframe}\boldsymbol{v}_{\imuframe}$ is the velocity of the robot in ${\imuframe}$ frame, 
	$\boldsymbol{\xi}_j$ is the $j$th state of the space-based sliding window,
	and $\boldsymbol{c}$ is the control vector obtained by column straightening of the $4\times 4$ control net $\mathbf{C}$. 
	
\begin{figure}[t!]
		\vspace{0.1cm}
		\centering
		\includegraphics[width=0.9\linewidth]{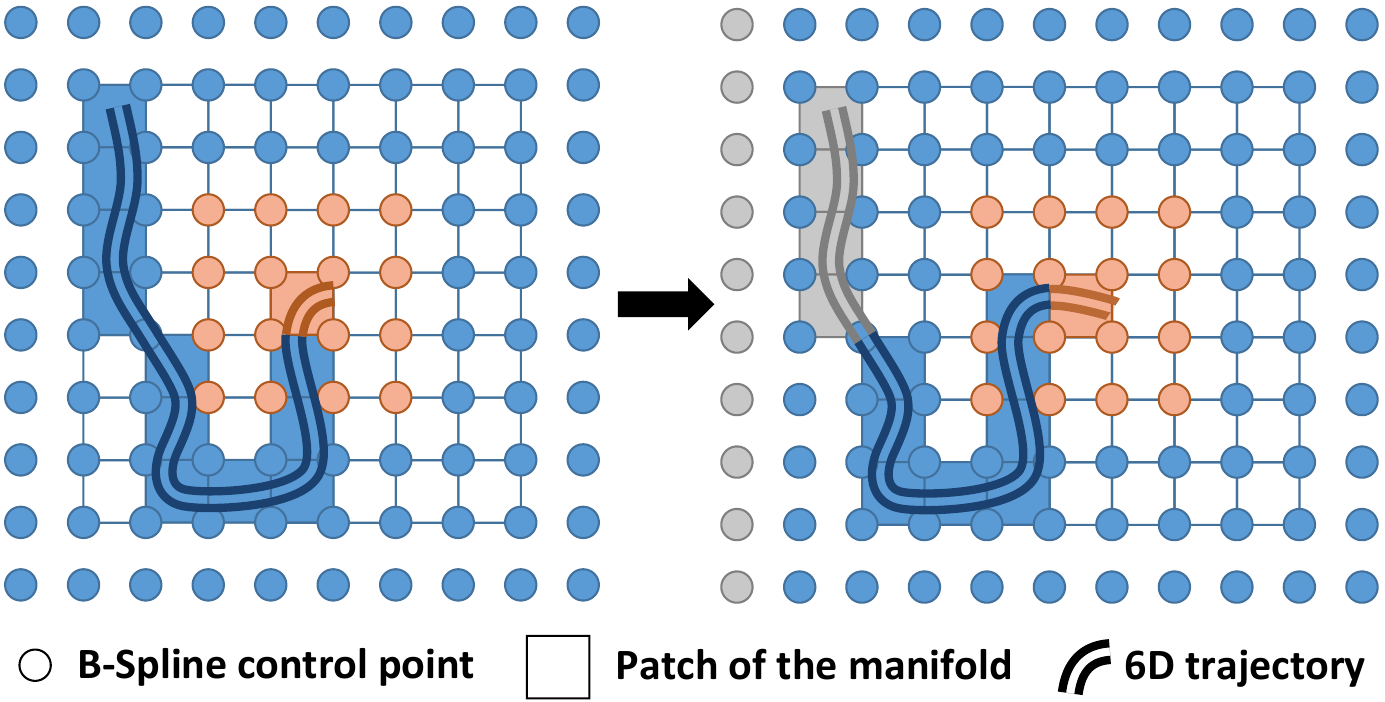}
		\caption{
			The demonstration of space-based sliding-window update strategy while the vehicle is moving in the rightward direction. Yellow represents active control points and patches of the manifold, blue represents static control points and patches, and gray represents marginalized control points and patches.}
		\label{fig:strategy}
		\vspace{-0.8cm}
	\end{figure}

	This work utilizes a uniform dual cubic B-spline manifold to characterize the ground on which the vehicle travels. Its knot vector has a constant interval $d_m$. All circles in Fig.~\ref{fig:strategy} demonstrate the control points. In practice, we only optimize the states associated with the patch of the B-spline manifold where the vehicle is currently traveling. The trajectories ($\mathcal{X}_S$, yellow lines) and control net ($\mathbf{C}$, $4\times4$ yellow circles) in Fig.~\ref{fig:strategy} represent the currently optimized states.
	\vspace{-0.1cm}
	\subsection{Process model}
	\label{Process_Model}
	In this work, we leverage the neural-processed IMU measurements to drive the state estimation system, whose full process model is given as follows:
	\begin{equation}
		\label{Process_model}
		\begin{aligned}
			{}^{\graframe}\dot{\boldsymbol{p}}_{\imuframe} &= 
			{}_{\imuframe}^{\graframe} \mathbf{R} \, {}^{\imuframe}\boldsymbol{v}_{\imuframe}, \\
			{}^{\imuframe}\dot{\boldsymbol{v}}_{\imuframe} &=
			{}^{\imuframe}{\accvel} - {_{\imuframe}^{\graframe}}\mathbf{R}^\top {^{\graframe}}\boldsymbol{\mathrm{g}} - 
			{}^{\imuframe}{\rotvel} \times {}^{\imuframe}\boldsymbol{v}_{\imuframe}, \\
			{}^{\graframe}_{\imuframe}{\dot{\psi}} &=
			\begin{bmatrix}
				-s_1& -s_2 & 1
			\end{bmatrix}  {}^{\imuframe}{\rotvel}, \\
			{}^{\graframe}_{\imuframe}\boldsymbol{\dot{s}} &= \frac{1}{2}
			\begin{bmatrix}
				-2 s_1 s_2 		  & s_1^2 - s_2^2 + 1 &  2 s_2 \\
				s_1^2 - s_2^2 - 1 &  2 s_1 s_2		  & -2 s_1 \\
			\end{bmatrix} {}^{\imuframe}{\rotvel},
		\end{aligned}
	\end{equation}
	where 
	the rotation $_{\imuframe}^{\graframe} \mathbf{R} = {}_{\imuframe}^{\graframe}\mathbf{R}_\psi {}_{\imuframe}^{\graframe} \mathbf{R}_\phi$ is expressed as a yaw-independent form described in Sec.~\ref{yaw_independent},
	and
	\vspace{-0.1cm}
	\begin{equation}
		\begin{aligned}
			{^{\imuframe}}{\rotvel} &= 
			{^{\imuframe}}{\widetilde{\rotvel}}-{^{\imuframe}}\widehat{\bias}_{\rotvel}-\ngvel, \quad
			{\ngvel} \sim \mathcal{N}(0,\widehat{\Cov}_{\rotvel}^2), \\
			{^{\imuframe}}{\accvel} &= 
			{^{\imuframe}}{\widetilde{\accvel}}-{^{\imuframe}}\widehat{\bias}_{\accvel}-\navel, \quad~
			{\navel} \sim \mathcal{N}(0,\widehat{\Cov}_{\accvel}^2).
		\end{aligned}
	\end{equation}
	In order to improve the efficiency of tuning the covariance of IMU measurements ${^{\imuframe}}\widehat{\bias}_{\rotvel}$ and ${^{\imuframe}}\widehat{\bias}_{\accvel}$ during sensor fusion and to exploit the consistency of the IMU bias and covariance of the network output, we can use the outputs of \textit{IMU De-Bias Net} to derive the IMU covariance as follows:
	\begin{equation}
		\begin{aligned}
			\label{thm1}
			\widehat{\Cov}_{\accvel}^2 = 
			\frac{1}{n\Delta t^2} \widehat{\Cov}_{\boldsymbol{v}_{i,i+n}}^2, \quad
			\widehat{\Cov}_{\rotvel}^2 =
			\frac{1}{n\Delta t^2} \widehat{\Cov}_{\boldsymbol{q}_{i,i+n}}^2.
		\end{aligned}
	\end{equation}
	The details of formula derivation can be found in Appendix~\ref{Proof}.
	Therefore, the differential equations can be discretized as follows:
	\vspace{-0.1cm}
	\begin{equation}
		\label{discrete_process_model}
		\begin{aligned}
			\mathcal{X}_{^{\imuframe_{k+1}}}
			&= \mathbf{F}(\mathcal{X}_{^{\imuframe_{k}}},\boldsymbol{u}_{k}) \\
			&= \mathbf{F}_{\imuframe} \mathcal{X}_{^{\imuframe_{k}}} + 
			\mathbf{F}_{\boldsymbol{n}} \boldsymbol{n}_{k},
		\end{aligned}
	\end{equation}
	where
	\vspace{-0.1cm}
	\begin{equation} 
		\begin{aligned}
			\mathcal{X}_{^{\imuframe_{k}}} &=
			\begin{bmatrix}
				{{}^{\graframe}} \boldsymbol{p}_{\imuframe_{k}}&
				{{}^{\imuframe_{k}}} \boldsymbol{v} &
				{{}^{\graframe}_{\imuframe_{k}}} \psi &
				{{}^{\graframe}_{\imuframe_{k}}} \boldsymbol{s}
			\end{bmatrix}^\top, \\[1em]
		\end{aligned}
	\end{equation}
	\vspace{-0.1cm}
	\begin{equation} 
		\begin{aligned}
			\boldsymbol{u}_{k} &=
			\begin{bmatrix}
				{{}^{\imuframe_k}}{\widehat{\rotvel}}-\ngvel \\[0.5em]
				{{}^{\imuframe_k}}{\widehat{\accvel}}-\navel
			\end{bmatrix}=
			\begin{bmatrix}
				{{}^{\imuframe}}{\widetilde{\rotvel}}-{{}^{\imuframe}}\widehat{\bias}_{\rotvel}-\ngvel \\[0.5em]
				{{}^{\imuframe}}{\widetilde{\accvel}}-{{}^{\imuframe}}\widehat{\bias}_{\accvel}-\navel
			\end{bmatrix},
		\end{aligned}
	\end{equation}
	where $k$ and $k+1$ denote timestamps. Details of $\mathbf{F}_{\imuframe}$ and $\mathbf{F}_{\boldsymbol{n}}$ can be found in Appendix~\ref{jacob_fx}
	\vspace{-0.1cm}
	\subsection{Observation model}
	\vspace{-0.2cm}
	There are two kinds of observation in the proposed system, one for the network-corrected velocity and the other for the manifold soft constraint.	
	\label{Observation_Model}
	\subsubsection{network-corrected velocity} 
	We could obtain triaxial network-corrected velocity of IMU through the~\textit{Wheel Encoder Net} in Sec.~\ref{Wheel Odometer Net}:
	\begin{equation}
		\label{observation_v}
		\begin{aligned}
			\boldsymbol{h}_{\boldsymbol{v}}(\mathcal{X}) 
			={}^{\imuframe}\boldsymbol{v}_{\imuframe}
			={}^{\imuframe}\boldsymbol{\widehat{v}}_{\imuframe}  + \boldsymbol{n}_{\boldsymbol{v}},
		\end{aligned}
	\end{equation}
	where
	$\boldsymbol{n}_{\boldsymbol{v}} \sim \mathcal{N}(0,\widehat{\Cov}_{\boldsymbol{v}}^2)$, 
	${}^{\imuframe}\boldsymbol{\widehat{v}}_{\imuframe}$ and $\widehat{\Cov}_{\boldsymbol{v}}^2$ can be obtained from the~\textit{Wheel Encoder Net}.

	\subsubsection{Manifold soft constraint}
	The IMU motion system is often affected by noise, which can result in inaccurate state estimation. To address this issue, we use soft constraints based on the assumption of a dual cubic B-spline manifold, as stated in Eq.~\eqref{manifold_constraint_concise}, to improve performance. Due to the sharing of control points among neighboring patches in the cubic B-spline manifold, each $4 \times 4$ control net influences the surrounding $7 \times 7$ patches. Therefore, we take all the soft constraints within these $7 \times 7$ patches into account.	
		As shown in the left subfigure of Fig.~\ref{fig:strategy}, there are $7\times 7$ patches of the B-spline manifold centered around the ground robot. The $7\times 7$ patches have two parts. 
	
	The first part represents the current patch $\mathcal{M}_{g_x,g_y}$ (active patch), which corresponds to the most intermediate ground that the robot currently stays on. In the active patch, we can get observation equations:
	\begin{equation}
		\label{observation_active}
		\begin{aligned}
			&\boldsymbol{h}_{g_x,g_y}(\mathcal{X})=
			\begin{bmatrix}
				\boldsymbol{h}_{g_x,g_y,\boldsymbol{\xi}_1}(\mathcal{X}) \\
				\vdots \\
				\boldsymbol{h}_{g_x,g_y,\boldsymbol{\xi}_{n_{g_x,g_y}}}(\mathcal{X})
			\end{bmatrix},
			\forall~\boldsymbol{\xi}_{j \leq n} \in \mathcal{X}_S \subseteq \mathcal{M}_{g_x,g_y}, \\
			&\boldsymbol{h}_{g_x,g_y,\boldsymbol{\xi}}(\mathcal{X}) = \boldsymbol{0}_{3 \times 1} = \boldsymbol{n}_{\boldsymbol{\mathcal{M}}} + \\&
			\begin{small}
				\begin{bmatrix}
					(\boldsymbol{y}_w \mathbf{K}_y \mathbf{B} \boxtimes
					\boldsymbol{x}_w  \mathbf{K}_x \mathbf{B}) Vec(\mathbf{C}_{g_x,g_y}) - z_w \\
					(\boldsymbol{\partial y}_w  \mathbf{K}_y \mathbf{B} \boxtimes
					\boldsymbol{x}_w  \mathbf{K}_x \mathbf{B}) Vec(\mathbf{C}_{g_x,g_y}) + 2 \frac{s_1 \sin(\psi) + s_2 \cos(\psi)}{1-s_1^2-s_2^2}\\ 
					(\boldsymbol{y}_w  \mathbf{K}_y \mathbf{B} \boxtimes
					\boldsymbol{\partial x}_w  \mathbf{K}_x \mathbf{B}) Vec(\mathbf{C}_{g_x,g_y}) + 2 \frac{s_1 \cos(\psi) - s_2 \sin(\psi)}{1-s_1^2-s_2^2}
				\end{bmatrix},
			\end{small}
		\end{aligned}
	\end{equation}
	where $
	\boldsymbol{n}_{\boldsymbol{\mathcal{M}}} \sim \mathcal{N}(0,\Cov_{\boldsymbol{\mathcal{M}}}^2)
	$ and $\mathbf{C}_{g_x,g_y}$ (active control net) is responsible for determining the active patch $\mathcal{M}_{g_x,g_y}$. The poses located in $\mathcal{M}_{g_x,g_y}$ are referred to as active poses.
	
	The other part represents the other 48 patches $\mathcal{M}_{u \neq g_x ,v \neq g_y}$ (static patches), which correspond to all the patches except the one on which the robot currently stays on. These static patches are determined through a combination of partial $\mathbf{C}_{g_x,g_y}$ and partial $\mathbf{C}_{u\neq g_x,v\neq g_y}$ (static control net). The poses located in $\mathcal{M}_{u \neq g_x ,v \neq g_y}$ are referred to as static poses. In these static patches, we can obtain observation equations:
	\begin{equation}
		\label{observation_static}
		\begin{aligned}
			&\boldsymbol{h}_{u,v}(\mathcal{X}) =
			\begin{bmatrix}
				\boldsymbol{h}_{u,v,\boldsymbol{\xi}_1}(\mathcal{X}) \\
				\vdots \\
				\boldsymbol{h}_{u,v,\boldsymbol{\xi}_{n_{u,v}}}(\mathcal{X})
			\end{bmatrix}, 
			\forall~\boldsymbol{\xi}_{j \leq n} \subseteq \mathcal{M}_{u \neq g_x ,v \neq g_y}, \\
			&\boldsymbol{h}_{u,v,\boldsymbol{\xi}}(\mathcal{X}) = \boldsymbol{0}_{3 \times 1} = \boldsymbol{n}_{\boldsymbol{\mathcal{M}}} + \\&
			{\small
				\begin{bmatrix}
					(\boldsymbol{y}_w \mathbf{K}_y \mathbf{B} \boxtimes
					\boldsymbol{x}_w  \mathbf{K}_x \mathbf{B}) \widetilde{Vec}(\mathbf{C}_{u,v}) - z_w \\
					(\boldsymbol{\partial y}_w  \mathbf{K}_y \mathbf{B} \boxtimes
					\boldsymbol{x}_w  \mathbf{K}_x \mathbf{B}) \widetilde{Vec}(\mathbf{C}_{u,v}) + 2 \frac{s_1 \sin(\psi) + s_2 \cos(\psi)}{1-s_1^2-s_2^2}\\ 
					(\boldsymbol{y}_w  \mathbf{K}_y \mathbf{B} \boxtimes
					\boldsymbol{\partial x}_w  \mathbf{K}_x \mathbf{B}) \widetilde{Vec}(\mathbf{C}_{u,v}) + 2 \frac{s_1 \cos(\psi) - s_2 \sin(\psi)}{1-s_1^2-s_2^2}
			\end{bmatrix}}, \\ 
			\small
			&\widetilde{Vec}(\mathbf{C}_{u,v}) = \\
			&\mathbf{M}_{S1}(u,v) Vec(\mathbf{C}_{g_x,g_y})+ 
			\mathbf{M}_{S2}(u,v) Vec(\mathbf{C}_{u\neq g_x,v\neq g_y}),
		\end{aligned}
	\end{equation}
	where $\widetilde{Vec}(\mathbf{C}_{u,v})$ is a vector that can be separated into the control points that will be optimized $(Vec(\mathbf{C}_{g_x,g_y}))$ and those that will not be optimized $(Vec(\mathbf{C}_{u\neq g_x,v\neq g_y}))$. This separation is achieved through two selection matrices, $\mathbf{M}_{S1}(u,v)$ and $\mathbf{M}_{S2}(u,v)$. All vectors and matrices in Eq.~\eqref{observation_active} and Eq.~\eqref{observation_static} are defined in the same way as in Sec.~\ref{Ground_manifold_constraint}. 
	\begin{algorithm}[t]
		\caption{Space-based sliding-window fusion.}
		\label{alg:ekf}
		\begin{algorithmic}[1]
				\State State: $\mathcal{X} = 
				\begin{bmatrix}
					\mathcal{X}_I {}^\top &
					\mathcal{X}_S {}^\top &
					\boldsymbol{c} {}^\top
				\end{bmatrix} {}^\top$;
				\State $\mathcal{X}_I$ initialization (Sec.\ref{ekf:init});
				\State $\boldsymbol{c}\_Initialized = False$ 
				\State $\mathcal{X}_S = \emptyset$
				\While{not at the end}
				\If{robot runs longer than a distance $d_s$}
				\State State augmentation (Sec.\ref{ekf:aug});
				\Else
				\State State propagation (Sec.\ref{ekf:prop}); 
				\EndIf
				\If{get network-corrected velocity}
				\State State update (Sec.\ref{ekf:update});
				\ElsIf{robot runs beyond the current patch}
				\If{$\boldsymbol{c}\_Initialized == False$}
				\State $\boldsymbol{c}$ initialization (Sec.\ref{ekf:init}); 
				\State $\boldsymbol{c}\_Initialized = True$
				\EndIf
				\State State update (Sec.\ref{ekf:update});
				\State Sliding and Marginalization (Sec.\ref{ekf:marg}); 
				\EndIf
				\EndWhile
		\end{algorithmic}
	\end{algorithm}
	
	Among the $7\times 7$ patches shown in the left subfigure of Fig.~\ref{fig:strategy}, only the active control net $\mathbf{C}_{g_x,g_y}$ and the active poses are estimated in one update process. 
	The static control net $\mathbf{C}_{u\neq g_x,v\neq g_y}$ and static poses are not optimized but utilized to construct the manifold soft constraints.
	By combining the soft constraints in each patch, we can obtain the observation equation with Gaussian white noise:
	\begin{equation}
		\begin{small}
			\setlength{\arraycolsep}{1.0pt}
			\label{observation_m}
			\begin{aligned}
				&\boldsymbol{h}_{\boldsymbol{\mathcal{M}}}(\mathcal{X}) = \\
				&Vec 
				\begin{pmatrix}
					\boldsymbol{h}_{g_x-3,g_y-3}(\mathcal{X}) & \cdots & 
					\boldsymbol{h}_{g_x-3,g_y}(\mathcal{X})   & \cdots & 
					\boldsymbol{h}_{g_x-3,g_y+3}(\mathcal{X}) \\
					\vdots & \ddots & \vdots & \ddots & \vdots \\
					\boldsymbol{h}_{g_x,g_y-3}(\mathcal{X})   & \cdots & 
					\boldsymbol{h}_{g_x,g_y}(\mathcal{X})     & \cdots & 
					\boldsymbol{h}_{g_x,g_y+3}(\mathcal{X}) \\
					\vdots & \ddots & \vdots & \ddots & \vdots \\
					\boldsymbol{h}_{g_x+3,g_y-3}(\mathcal{X}) & \cdots & 
					\boldsymbol{h}_{g_x+3,g_y}(\mathcal{X})   & \cdots & 
					\boldsymbol{h}_{g_x+3,g_y+3}(\mathcal{X}) \\
				\end{pmatrix},
			\end{aligned}
		\end{small}
	\end{equation}
	where each $\boldsymbol{h}_{u,v}(\mathcal{X})$ represents $3\times n_{u,v}$ equations in one patch, where $u \in \{g_x, g_x\pm1, g_x\pm2, g_x\pm3 \} $, $v \in \{g_y, g_y\pm1, g_y\pm2, g_y\pm3\}$, and $n_{u,v}$ is the number of states $\boldsymbol{\xi}_j$ in each patch. 
	\subsection{Extended Kalman filtering process}
	The algorithm presented in Algorithm~\ref{alg:ekf} outlines the filtering process. The specific details are elaborated in the subsequent sections.
	\subsubsection{Initialization}
	\label{ekf:init}
	The proposed filter framework initializes the IMU state $\mathcal{X}_I$ as the ground truth, and covariances of these states are initialized as shown in TABLE~\ref{intial_value_cov} based on experiments. Besides, other system parameters are pre-defined as presented in TABLE~\ref{exp_params}.
		The space-based sliding-window states $\mathcal{X}_S$ are initially empty, and states will be added or marginalized according to the procedures described in Section~\ref{ekf:aug} and Section~\ref{ekf:marg}.
		When the robot runs beyond one patch of the B-spline manifold for the first time, the control vector $\boldsymbol{c}$ is initialized by the least squares method as follows:
	\begin{equation}
		\label{control_mesh_initial}
		\begin{aligned}
			\boldsymbol{c} &= (\mathbf{A}^\top \mathbf{A})^{-1} \mathbf{A}^\top \boldsymbol{b},\quad
			\forall~\boldsymbol{\xi}_{j \leq n} \in \mathcal{X}_S \subseteq \mathcal{M}_{g_x,g_y},\\
			\mathbf{A} &=
			\begin{bmatrix}
				\boldsymbol{A}_{\boldsymbol{\xi}_1},
				\boldsymbol{A}_{\boldsymbol{\xi}_2},
				\hdots,
				\boldsymbol{A}_{\boldsymbol{\xi}_n}
			\end{bmatrix}^\top, \quad \ 
			\boldsymbol{b} =
			\begin{bmatrix}
				\boldsymbol{b}_{\boldsymbol{\xi}_1},
				\boldsymbol{b}_{\boldsymbol{\xi}_2},
				\hdots
				\boldsymbol{b}_{\boldsymbol{\xi}_n}
			\end{bmatrix}^\top,\\
			\boldsymbol{A_{\xi}} &=
			\begin{bmatrix}
				\boldsymbol{y}_w \mathbf{K}_y \mathbf{B} \otimes
				\boldsymbol{x}_w \mathbf{K}_x \mathbf{B}\\
				\boldsymbol{\partial y}_w \mathbf{K}_y \mathbf{B} \otimes
				\boldsymbol{x}_w \mathbf{K}_x \mathbf{B}\\ 
				\boldsymbol{y}_w \mathbf{K}_y \mathbf{B} \otimes
				\boldsymbol{\partial x}_w \mathbf{K}_x \mathbf{B}
			\end{bmatrix},
			\boldsymbol{b_{\xi}} =
			\begin{bmatrix}
				z_w\\
				-2 \frac{s_1 \sin(\psi) + s_2 \cos(\psi)}{1-s_1^2-s_2^2}\\
				2 \frac{s_1 \cos(\psi) - s_2 \sin(\psi)}{1-s_1^2-s_2^2}
			\end{bmatrix}
		\end{aligned}
	\end{equation}
	where $\boldsymbol{\xi}_n$ represents the $n$th state included in $\mathcal{X}_S$ through the state augment process (Sec.~\ref{ekf:aug}).
	
	\subsubsection{State propagation}
	\label{ekf:prop}
	We can obtain the discrete state propagation equations from the derivations in the Sec.~\ref{Process_Model}:
	\begin{equation}
		\label{eq:prop}
		\begin{aligned}
			\mathcal{X}_{^{\imuframe_{k+1}}} &= 
			\mathbf{F}(\mathcal{X}_{^{\imuframe_{k}}},\boldsymbol{u}_{k}),\\
			\mathbf{P}_{k+1} &= \mathbf{A}_{k} \mathbf{P}_{k} \mathbf{A}_{k}^\top +
			\mathbf{B}_{k} \mathbf{W} \mathbf{B}_{k}^\top, \\
			\mathbf{A}_{k} = 
			\begin{bmatrix}
				\mathbf{F}_\imuframe & \mathbf{0} \\
				\mathbf{0} & \mathbf{I}
			\end{bmatrix}&, \quad 
			\mathbf{B}_{k} = 
			\begin{bmatrix}
				\mathbf{F}_{\boldsymbol{n}} \\
				\mathbf{0}
			\end{bmatrix}, \quad
			\mathbf{W} = 
			\begin{bmatrix}
				\Cov_{\accvel}^2 & \mathbf{0} \\
				\mathbf{0} & \Cov_{\rotvel}^2
			\end{bmatrix},
		\end{aligned}
	\end{equation}
	where $k$ and $k+1$ represent timestamps.
	\begin{table}[t]
		\vspace{0cm}
		\centering
		\caption{States Initialization}
		\scalebox{0.9}{
			\begin{tabular}{cccccccc}
				\toprule
				& \makecell[c]{${^{\graframe}}\boldsymbol{p}{_{\imuframe}}$}
				& \makecell[c]{${^{\imuframe}}\boldsymbol{v}{_{\imuframe}}$}
				& \makecell[c]{${^{\graframe}_{\imuframe}{\psi}}$}
				& \makecell[c]{${^{\graframe}_{\imuframe}\boldsymbol{s}}$}
				& \makecell[c]{${\boldsymbol{c}}$} \\ 
				\toprule
				Value
				& ${^{\graframe}}\boldsymbol{p}{_{\imuframe_0}}$	
				& ${^{\imuframe_0}}\boldsymbol{v}{_{\imuframe_0}}$   
				& ${^{\graframe}_{\imuframe_0}{\psi}}$ 
				& ${^{\graframe}_{\imuframe_0}\boldsymbol{s}}$ 
				& ${\text{LSM}}$  \\
				\textit{Diag}(cov)  	
				& $1\mathrm{e}{-8}$   
				& $1\mathrm{e}{-8}$    	
				& $1\mathrm{e}{-8}$      
				& $1\mathrm{e}{-8}$ 
				& $1\mathrm{e}{-2}$ \\
				\bottomrule
		\end{tabular}}
		\label{intial_value_cov}
		\vspace{-0.3cm}
	\end{table}
	
	\begin{table}[t]
			\vspace{0cm}
			\centering
			\caption{Experiment Parameters}
			\scalebox{0.9}{
					\begin{tabular}{cccccccccc}
						\toprule
						\makecell[c]{Parameter}
						& \makecell[c]{Value} 	\\ 
						\toprule
						Interval of knot vector  ${d_m}$
						& ${5m}$ \\
						Covariance of manifold soft constraints 
						${\boldsymbol{n}_{\boldsymbol{\mathcal{M}}}}$
						& $1\mathrm{e}{1}$ \\
						Distance for state augmentation $d_s$ 
						& ${0.1m}$ \\
						\bottomrule
			\end{tabular}}
			\label{exp_params}
			\vspace{-0.6cm}
	\end{table}
	\subsubsection{State augmentation}
	\label{ekf:aug}
	In the proposed system, the dimension of the space-based sliding-window states $\mathcal{X}_{S}$ grows as the ground robot runs longer than a fixed distance (parameter $d_s$). During an augmentation step, cloning is performed together with propagation in one step. 
	\begin{equation}
		\label{eq:aug}
		\begin{aligned}
			\mathcal{X}_{^{\imuframe_{k+1}}} & 
			=\mathbf{F}(\mathcal{X}_{^{\imuframe_{k}}},\boldsymbol{u}_{k}),\\
			\mathcal{X}_{^{S_{k+1}}} &
			= [ \mathcal{X}_{^{S_{k}}}, \boldsymbol{\xi}_{n_{k+1}} ], \\
			\mathbf{P}_{k+1} &= 
			\mathbf{\bar{A}}_{k} \mathbf{P}_{k} \mathbf{\bar{A}}_{k}^\top +
			\mathbf{\bar{B}}_{k} \mathbf{W} \mathbf{\bar{B}}_{k}^\top, \\
			\boldsymbol{\xi}_{n_{k+1}} &
			= 
				\begin{bmatrix}
					^{\graframe}\boldsymbol{p}_{\imuframe_{k+1}}^\top & 
					^{\graframe}_{\imuframe_{k+1}}{\psi} &
					^{\graframe}_{\imuframe_{k+1}}\boldsymbol{s}^\top
				\end{bmatrix} {}^\top \\
			\mathbf{\bar{A}}_{k} &
			= 
				\begin{bmatrix}
					\mathbf{F}_\imuframe & \mathbf{0}  \\
					\mathbf{0} & \mathbf{I} \\
					\mathbf{F}_\imuframe^{\boldsymbol{\xi_{k}}} & \mathbf{0} 
				\end{bmatrix}, \quad 
				\mathbf{\bar{B}}_{k} = 
				\begin{bmatrix}
					\mathbf{F}_{\boldsymbol{n}} \\
					\mathbf{0} \\
					\mathbf{F}_{\boldsymbol{n}}^{\boldsymbol{\xi_{k}}}, 
				\end{bmatrix},
		\end{aligned}
	\end{equation}
	where $k$ and $k+1$ represent timestamps. $\boldsymbol{\xi}_{n_{k+1}}$ represents the $(n+1)$th $\boldsymbol{\xi}$ in the $\mathcal{X}_{S}$, which contains the rotation and position in the $\imuframe$ frame at the timestamp $k+1$. $\mathbf{\bar{A}}_{k}$ and $\mathbf{\bar{B}}_{k}$ involve both copying and the propagation of augmented and current states.  $\mathbf{F}_\imuframe^{\xi_{k}}$ and $\mathbf{F}_{\boldsymbol{n}}^{\xi_{k}}$ are partial propagation matrics for rotation and position, which only maintain the Jacobian related to rotation and position. After the state augmentation step, the dimension of the space-based sliding-window states $\mathcal{X}_{S}$ increases by $6$. 
	\subsubsection{Observation update}
	\label{ekf:update}
	There are two observation updates in the proposed EKF. One is a network-corrected velocity in the IMU frame ($\imuframe$). 
	And the other is a soft constraint of the ground manifold, which is used to update the state when the ground robot runs beyond the boundary of the current patch of the B-spline manifold. We use the conventional extended Kalman filtering process for observation updates:
	\begin{equation}
		\label{ekf-obs}
		\begin{aligned}
			\mathbf{K} &= \mathbf{P} \mathbf{H}^\top
			\left(
			\mathbf{H}\mathbf{P}\mathbf{H}^\top + \mathbf{S}
			\right)^{-1},\\
			\mathcal{X} &\leftarrow \mathcal{X} + 
			\mathbf{K}\big(\boldsymbol{z}-\boldsymbol{h}(\mathcal{X})\big),\\
			\mathbf{P} &\leftarrow (\mathbf{I}-\mathbf{K}\mathbf{H})\mathbf{P}(\mathbf{I}-\mathbf{K}\mathbf{H})^\top + \mathbf{K} \mathbf{S} \mathbf{K}^\top, 
		\end{aligned}
	\end{equation}
	where $\mathbf{H}$, $\mathbf{S}$, $\boldsymbol{h}(\mathcal{X})$ and $\boldsymbol{z}$ denote the Jacobian matrix of the observation model, observation covariance, observation model and observation value provided by Sec.~\ref{Observation_Model}, respectively. The Jacobian matrix $\mathbf{H}$ corresponding to each observation model ($\mathbf{H}_{\boldsymbol{v}}$, $\mathbf{H}_{\boldsymbol{\mathcal{M}_1}}$, $\mathbf{H}_{\boldsymbol{\mathcal{M}_2}}$) can be found in Appendix~\ref{jacob_obs}. The observation covariance $\mathbf{S}$ is $\Cov_{\boldsymbol{v}}^2$ when observations are derived from network-corrected velocity, $\Cov_{\boldsymbol{\mathcal{M}}}^2$ when observations stem from manifold soft constraints, or $\begin{bmatrix}
			\Cov_{\boldsymbol{v}}^2 & \mathbf{0} \\
			\mathbf{0} & \Cov_{\boldsymbol{\mathcal{M}}}^2
		\end{bmatrix}$ when both types of observations are present.	
	\subsubsection{Sliding and marginalization}
	\label{ekf:marg}
	In Fig.~\ref{fig:strategy}, when the ground robot travels to the next patch of the B-spline manifold, 
	the trajectory poses that are located in the previous patch and the control net that no longer shapes the previous patch are both removed from the active poses and added to the static poses. 
	Only when the static poses and static control net cannot affect the current 7$\times$7 patches that are centered around the robot, will they be marginalized, that is, dropped into the marginalized states as shown in the right subfigure of Fig.~\ref{fig:strategy}.
	\begin{figure*}[t]
		\vspace{0.0cm}
		\centering
		\includegraphics[width=\linewidth]{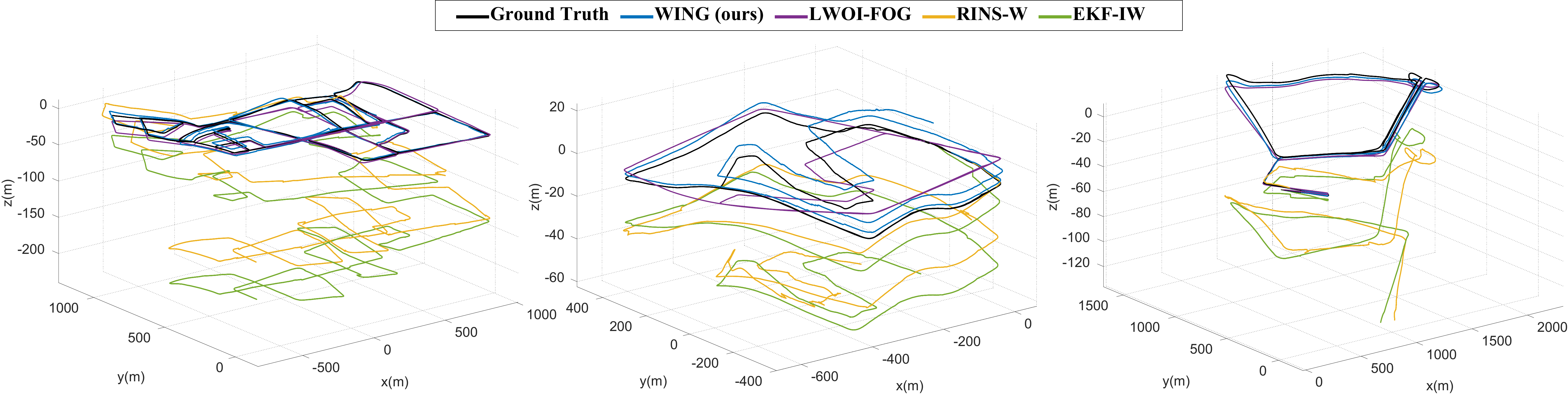}
		\caption{
			Demonstration of different algorithms for pose estimation.
			From left to right, there are 3D estimated trajectories of \textit{urban09}, \textit{urban15}, and \textit{urban17} of the KAIST Urban Dataset~\cite{jeong2019complex}.
		}
		\label{fig:traj}
		\vspace{-0.6cm}
	\end{figure*}
	\section{Experiments}
	To better evaluate the proposed method, we compare estimation accuracy with other approaches and perform ablation studies to demonstrate the performance of each module on KAIST Urban Dataset~\cite{jeong2019complex} and NCLT Dataset~\cite{carlevaris2016university}. 
		\subsection{Implementation details}
		\subsubsection{Datasets}
		The KAIST Urban Dataset~\cite{jeong2019complex} collected by ground vehicles in different cities in South Korea. We utilized the Xsens MTi-300 IMU and RLS LM13 wheel encoders as the sensors from the KAIST Urban Dataset~\cite{jeong2019complex}. All measurements captured by these sensors were recorded at a frequency of 100Hz.
		The NCLT Dataset~\cite{carlevaris2016university} contains data of exploring the campus, both indoors and outdoors. Microstrain GX3 IMU Segway’s wheel encoders from the NCLT Dataset~\cite{carlevaris2016university} are used in our experiments. For simpler implementation, we standardize the frequency of measurements from the NCLT Dataset~\cite{carlevaris2016university} to 100 Hz. We achieved this by applying nearest-neighbor interpolation to both the sensor measurements and the ground truth data in the NCLT Dataset~\cite{carlevaris2016university}.
		To obtain the ground truth velocity for training and evaluation purposes from both KAIST Urban Dataset~\cite{jeong2019complex} and NCLT Dataset~\cite{carlevaris2016university}, we fit the robot's position trajectory using a cubic polynomial curve and then differentiate it to calculate the corresponding velocity.
		\subsubsection{Training details}
		We utilize the KAIST Urban Dataset~\cite{jeong2019complex} to train and validate the \textit{IMU De-bias Net} and \textit{Wheel Encoder Net}. 
		Specifically, we randomly select sequences \textit{urban07}, \textit{urban09}, \textit{urban11}, \textit{urban13}, \textit{urban15}, and \textit{urban17} for testing, while the remaining sequences are used for training and validation.
		To evaluate the network's generalization capabilities, we incorporate sequences from the NCLT Dataset~\cite{carlevaris2016university} as well. We additionally include the first 10 minutes of the \textit{12/08/20}, \textit{12/12/01}, and \textit{13/04/05} sequences from the NCLT Dataset~\cite{carlevaris2016university} in our test set.	
		The proposed networks use the Adam optimizer to minimize the loss function with an initial learning rate of $10^{-4}$.
	
	\subsection{Pose comparison}
	\label{exp:benchmark}
	We select the advanced learning-based interoceptive-only odometry methods for comparison: RINS-W~\cite{brossard2019rins}, TLIO~\cite{tlio2020liu}, LWOI~\cite{brossard2019learning} with FOG measurements~(LWOI-FOG), and LWOI~\cite{brossard2019learning} with IMU's gyroscope measurements~(LWOI-IMU). 
	In addition, we performed a comparison with a traditional inertial-wheel odometry method (referred to as EKF-IW), based on an open-source implementation\footnote{$https://github.com/gaoxiang12/slam\_in\_autonomous\_driving$}. The EKF-IW fuses raw IMU and wheel encoder measurements within an EKF framework, which includes online IMU bias estimation. To simulate the offline IMU calibration, we initialize the IMU bias using the ground truth and IMU measurements when the vehicle is stationary.
	In order to compare the  estimation accuracy, we use EVO~\cite{grupp2017evo} to evaluate the RMSE of absolute translation error ($\textbf{ATE}$) and rotation error ($\textbf{ARE}$):
	\begin{itemize}
		\label{equ:metric}
		\item{\textbf{ATE}}~(\textrm{m})$:= \sqrt{\frac{1}{M}\sum_{i=1}^M 
			\|{^{\graframe}}\boldsymbol{p}_i -
			{^{\graframe}}\boldsymbol{\widehat{p}}_i
			\|^2_2}$,\\
		\item{\textbf{ARE}}~(\textrm{deg})$:= \sqrt{\frac{1}{M}\sum_{i=1}^M
			\|\mathrm{Log}(\boldsymbol{\widehat{q}}_i^\ast \otimes \boldsymbol{q}_i)
			\|^2_2}$,
	\end{itemize}
	where $M$ is the number of the estimated poses.
	
	The TABLE~\ref{tab:pose_comparison} shows the translation and rotation RMSE results of all competing algorithms. 
	These results demonstrate that our proposed method, WING, can achieve more accurate pose estimation than RINS-W~\cite{brossard2019rins}, TLIO~\cite{tlio2020liu}, LWOI-IMU~\cite{brossard2019learning} and EKF-IW. 	
	Despite the pose accuracy difference between our method and LWOI-FOG~\cite{brossard2019learning} is not too large, it is worth noting that LWOI-FOG~\cite{brossard2019learning} relies on the use of an expensive FOG to calculate the yaw angle of the vehicle. FOGs are expensive and not commonly available in commercial vehicles. However, our system achieves competitive performance by utilizing affordable and widely used IMU and wheel encoder.
		In contrast to the EKF-IW method, the \textit{Wheel Encoder Net} offers improved accuracy in velocity observations, enabling more reliable updates within the EKF framework. Besides, the \textit{Wheel Encoder Net} corrects IMU measurements based on historical inputs independently. However, in EKF-IW, the bias estimation can be significantly influenced by noise from both the IMU and wheel encoder. What's more, The learned covariance from \textit{IMU De-Bias Net} provides a dynamic uncertainty prediction. In the EKF-IW, however, the covariance values for IMU inputs, bias random walk, and velocity observations remain fixed.
	The 3D trajectories are drawn in Fig.~\ref{fig:traj}, where TLIO~\cite{tlio2020liu} and LWOI-IMU~\cite{brossard2019learning} are not plotted because of their divergent estimations. 
	These results demonstrate the benefits of using the ground manifold, as well as the bias and velocity estimation from neural networks.
	\begin{table}[t]
		\caption{Translation and Rotation evaluation in KAIST Urban Dataset~\cite{jeong2019complex} and NCLT Dataset~\cite{carlevaris2016university}. The \textbf{\textcolor{red}{best}} and \textbf{second best} results are separately marked in red bold and black bold, respectively.}
		\label{tab:pose_comparison}
		\vspace{0.0cm}
		\centering
		\resizebox{\linewidth}{!}{
			\begin{tabular}{cccccccc}
				\toprule
				& ~
				& \textbf{RINS-W} & \textbf{\makecell{LWOI\\-IMU}} & \textbf{\makecell{LWOI\\-FOG}} & \textbf{TLIO}& \textbf{\makecell{EKF\\-IW}} & \textbf{Ours}  \\
				\toprule
				KAIST
				& ATE   
				& 18.41	& 409.41 & \textbf{\textcolor{red}{5.17}} & 96.74 & 30.99 & \textbf{8.10} \\
				urban07
				& ARE 
				& 2.34	& 84.56	& 5.08 & 44.96 & \textbf{2.24}& \textbf{\textcolor{red}{1.47}}	 \\
				\toprule
				KAIST
				& ATE   
				& 214.88 & 2402.82	& \textbf{\textcolor{red}{19.31}} & 846.22 & 146.03& \textbf{22.25}  \\
				urban09
				& ARE
				& 10.85	& 100.54	& \textbf{4.23} & 114.24 & 4.58& \textbf{\textcolor{red}{2.11}}	 \\
				\toprule
				KAIST
				& ATE      
				& 113.90 & 156.29 & \textbf{70.79} & 4024.93 & 373.67& \textbf{\textcolor{red}{40.25}}  \\
				urban11
				& ARE
				& 4.58	& 3.77	& \textbf{2.58} & 45.53 & 3.62& \textbf{\textcolor{red}{1.50}} \\
				\toprule
				KAIST
				& ATE      
				& 449.29 & 624.22 & \textbf{15.59} & 425.82 & 35.24& \textbf{\textcolor{red}{14.58}}  \\
				urban13
				& ARE 
				& 28.37	& 86.17	& \textbf{2.51} & 59.723& 4.48& \textbf{\textcolor{red}{1.94}} \\
				\toprule
				KAIST
				& ATE      
				& 69.86	& 206.62 & \textbf{8.08} & 333.15 & 39.44& \textbf{\textcolor{red}{7.75}}	 \\
				urban15
				& ARE
				& 8.55	& 32.35	& 3.78 & 65.18 & \textbf{3.08} & \textbf{\textcolor{red}{1.11}}	 \\
				\toprule
				KAIST
				& ATE      
				& 86.10	& 1384.02 & \textbf{\textcolor{red}{11.52}} & 955.16 & 80.74& \textbf{12.64}	\\
				urban17
				& ARE 
				& 4.49	& 57.33	& \textbf{3.12} & 69.87 & 3.28& \textbf{\textcolor{red}{0.57}}	 \\
				\toprule
				NCLT
				& ATE      
				& 286.47 & 188.24 & \textbf{15.43} & 240.77 & 50.77 & \textbf{\textcolor{red}{12.35}} \\
				12/08/20
				& ARE 
				& 86.22 & 38.36 & \textbf{4.89} & 29.88 & 13.68 & \textbf{\textcolor{red}{2.91}} \\
				\toprule
				NCLT
				& ATE      
				& 203.91 & 114.98 & \textbf{12.49} & 163.92 & 54.74 & \textbf{\textcolor{red}{9.53}} \\
				12/12/01
				& ARE 
				& 54.40 & 38.72 & \textbf{4.79} & 31.01 & 19.07 & \textbf{\textcolor{red}{2.48}} \\
				\toprule
				NCLT
				& ATE      
				& 251.21 & 251.21 & \textbf{12.60} & 141.34 & 37.20 & \textbf{\textcolor{red}{8.22}} \\
				13/04/05
				& ARE 
				& 60.12 & 60.12 & \textbf{4.99} & 40.47 & 9.71 & \textbf{\textcolor{red}{3.87}} \\
				\bottomrule
		\end{tabular}}
		\vspace{-0.6cm}
	\end{table}
	\subsection{Evaluation of each module}
	\begin{figure}[t!]
		\centering
		\includegraphics[width=\linewidth]{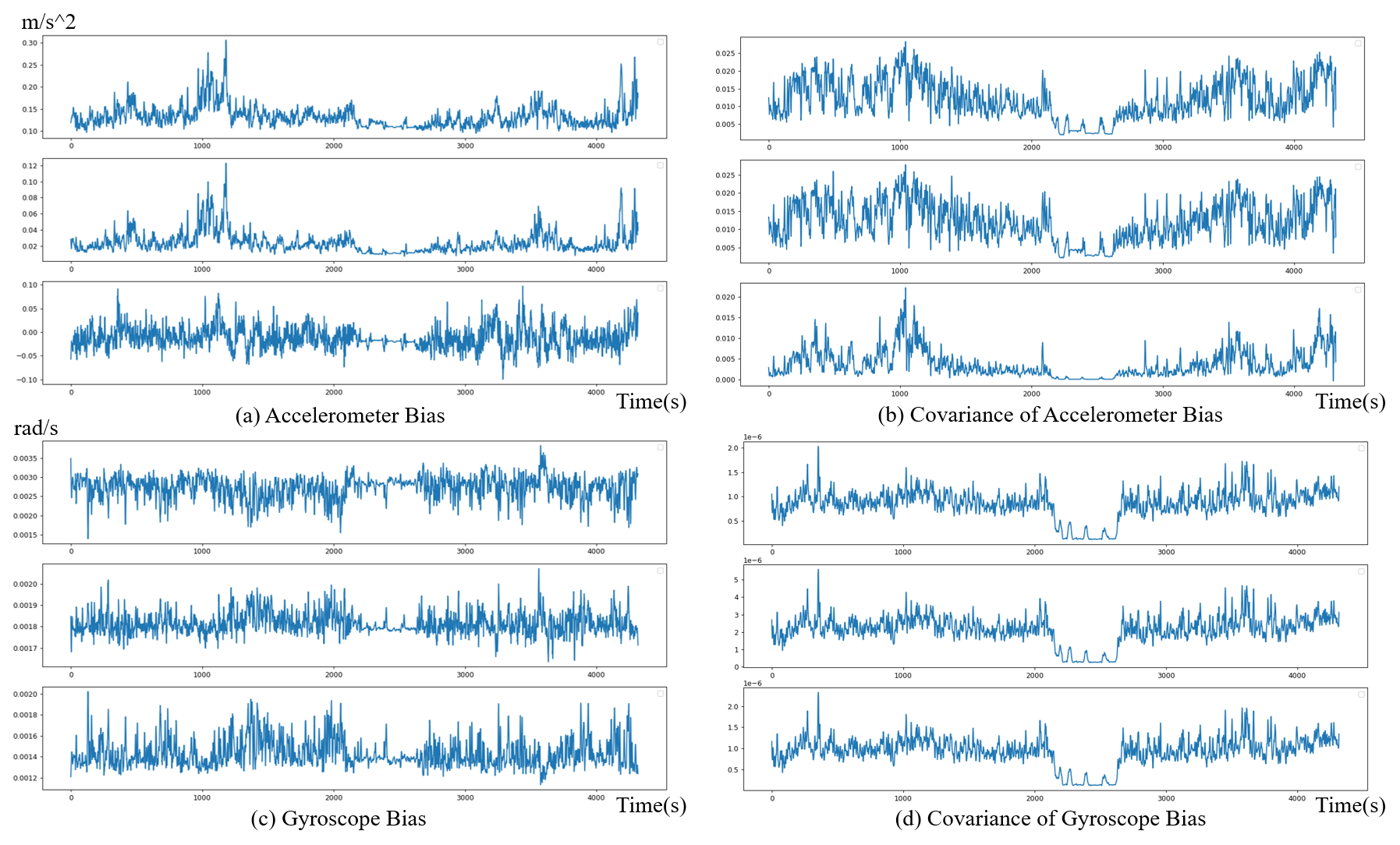}
		\caption{Predicted time-varying bias estimation and its corresponding covariance from \textit{IMU De-Bias Net} on the \textit{urban11} of the KAIST Urban Dataset ~\cite{jeong2019complex}.}
		\label{fig:time-vary-bias}
		\vspace{-0.2cm}
	\end{figure}
	\begin{figure}[t!]
		\vspace{-0.0cm}
		\centering
		\includegraphics[width=\linewidth]{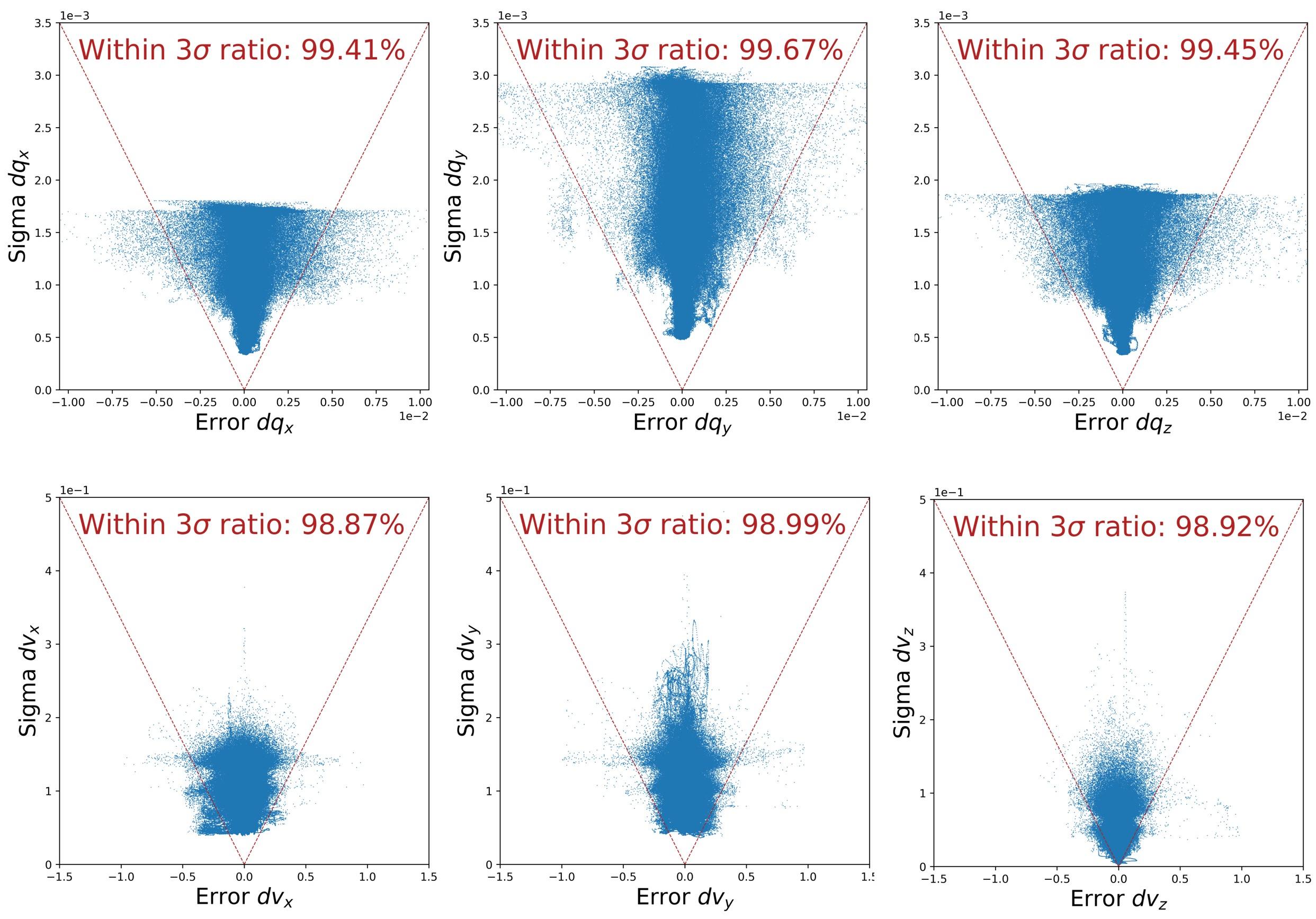}
		\caption{
			Standard deviation $\sigma$ against the errors in ${\boldsymbol{\widehat v}}_{i,i+n}$ (first row) and ${{\boldsymbol q}}_{i,i+n}$ (second row) along the x, y, and z axes. 
			Error $dv$ represents $\boldsymbol {v}_{i,i+n} - {\boldsymbol{\widehat v}}_{i,i+n}$, and error $dq$ denotes $\mathrm{Log} \big(
			\left({\boldsymbol{\widehat q}}_{i,i+n} \right)^* \otimes {\boldsymbol q}_{i,i+n} \big)$. The dashed red line represents the 3$\sigma$ region and the red text indicates the ratio of error points that fall within this 3$\sigma$ region.}
		\label{fig:sigma_debias}
		\vspace{-0.6cm}
	\end{figure}
	\subsubsection{\textit{IMU De-Bias Net}}
	As illustrated in Fig.~\ref{fig:time-vary-bias}, the \textit{IMU De-Bias Net} provides time-varying bias estimation along with its corresponding covariance. Since the performance of bias estimation has already been demonstrated in our previous work~\cite{zhang2022dido}, this study focuses solely on evaluating the performance of the new covariance branch. 
	To assess the consistency of the learned covariance, we gather samples from the complete test set. Fig.~\ref{fig:sigma_debias} presents the results of this analysis, which demonstrate that around 99\% of the error points fall within the 3$\sigma$ region. 
	These observations confirm that the outputs generated by the \textit{IMU De-Bias Net} exhibit high consistency. 
	Moreover, the results reveal that the covariance of $dq_y$ is significantly higher than that of $dq_x$ and $dq_z$. 
	This can be attributed to the fact that when a car is moving forward, its pitch angle is more susceptible to disturbances, leading to increased uncertainty in the estimation of $dq_y$. 
	\subsubsection{\textit{Wheel Encoder Net}}
	We evaluate the performance of \textit{Wheel Encoder Net} on the test set and present the results in TABLE~\ref{tab:v_body_comparison}. 
	These results demonstrate that \textit{Wheel Encoder Net} can significantly improve the accuracy of velocity estimation on the x-axis and y-axis. 
	However, the enhancement on the z-axis is not as significant. This is because vehicles are more likely to skid during turning, whereas vertical bouncing is relatively rare on smooth urban roads.
	
\begin{figure}[t]
		\vspace{0.0cm}
		\centering
		\includegraphics[width=\linewidth]{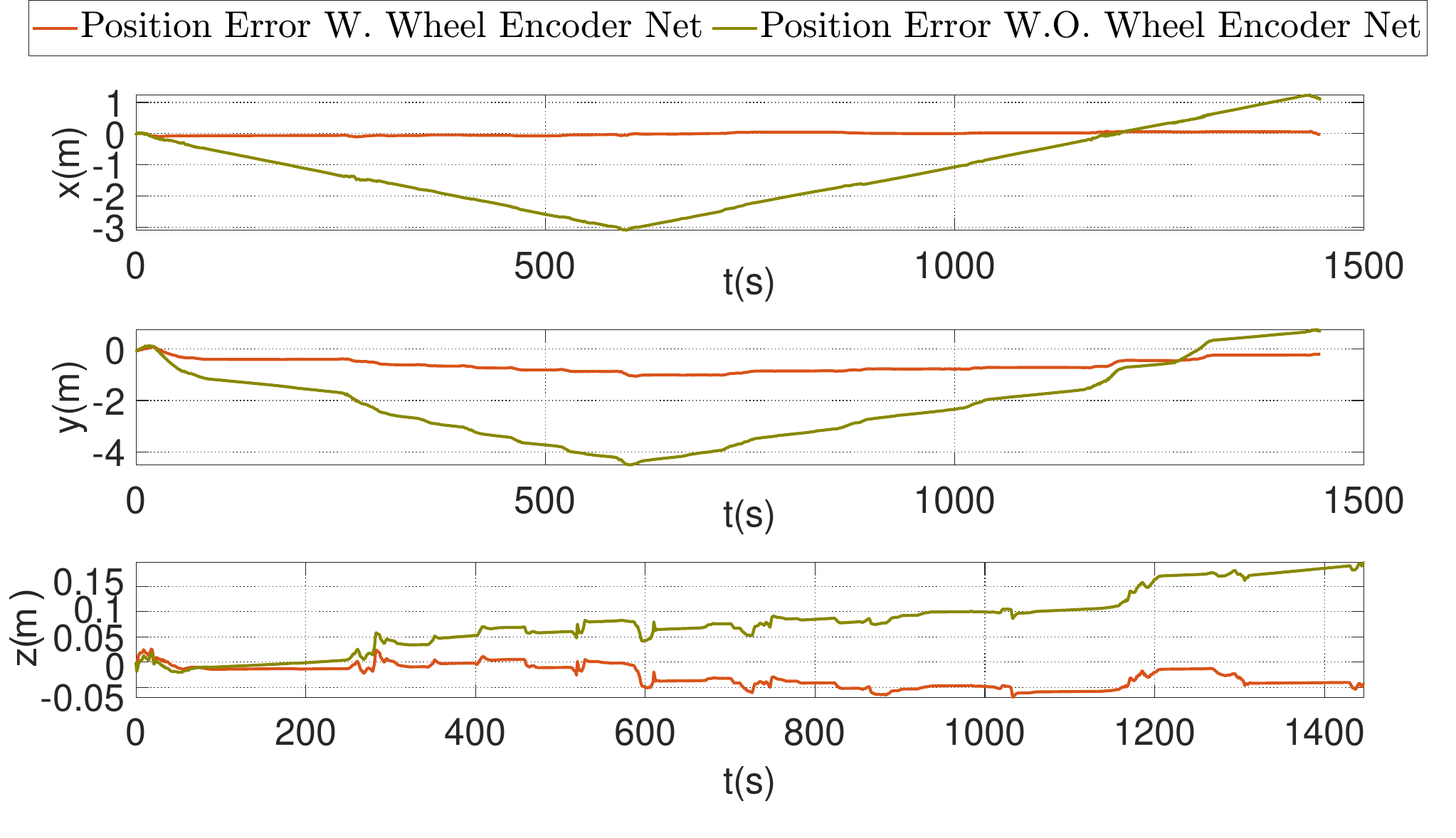}
		\caption{
			The error in position estimation resulting from integrating velocity when using the \textit{Wheel Encoder Net} compared to not using it on the \textit{urban13} of KAIST Urban Dataset ~\cite{jeong2019complex}.}
		\label{fig:p_inte_v}
		\vspace{-0.3cm}
	\end{figure}

	Fig.~\ref{fig:p_inte_v} illustrates an example of the improved position estimation achieved by utilizing the \textit{Wheel Encoder Net}. 
	We integrate the velocity obtained from both the \textit{Wheel Encoder Net} (${^{\imuframe}}{\boldsymbol{\widehat{v}}}_{{\imuframe}}$) and raw measurements(${}^{\imuframe}{\boldsymbol{\widetilde{v}}}_{\imuframe} = {{}_{\bodyframe}^{\imuframe}}\mathbf{R} ~ {}^{\bodyframe}{\boldsymbol{\widetilde{v}}}_{\bodyframe} - {}^{\imuframe}{\widetilde{\rotvel}} \times {}^{\imuframe}\boldsymbol{t}_{\bodyframe}-{}^{\imuframe}\dot{\boldsymbol{t}}_{\bodyframe}$) using the ground truth attitude. The ${}_{\bodyframe}^{\imuframe}\mathbf{R}$ and ${}^{\imuframe}\boldsymbol{t}_{\bodyframe}$ are obtained from the data sheet.
	As depicted in Fig.~\ref{fig:p_inte_v}, utilizing the \textit{Wheel Encoder Net} leads to more accurate position estimation.
	
\begin{figure}[h]
		\vspace{0.1cm}
		\centering
		\includegraphics[width=\linewidth]{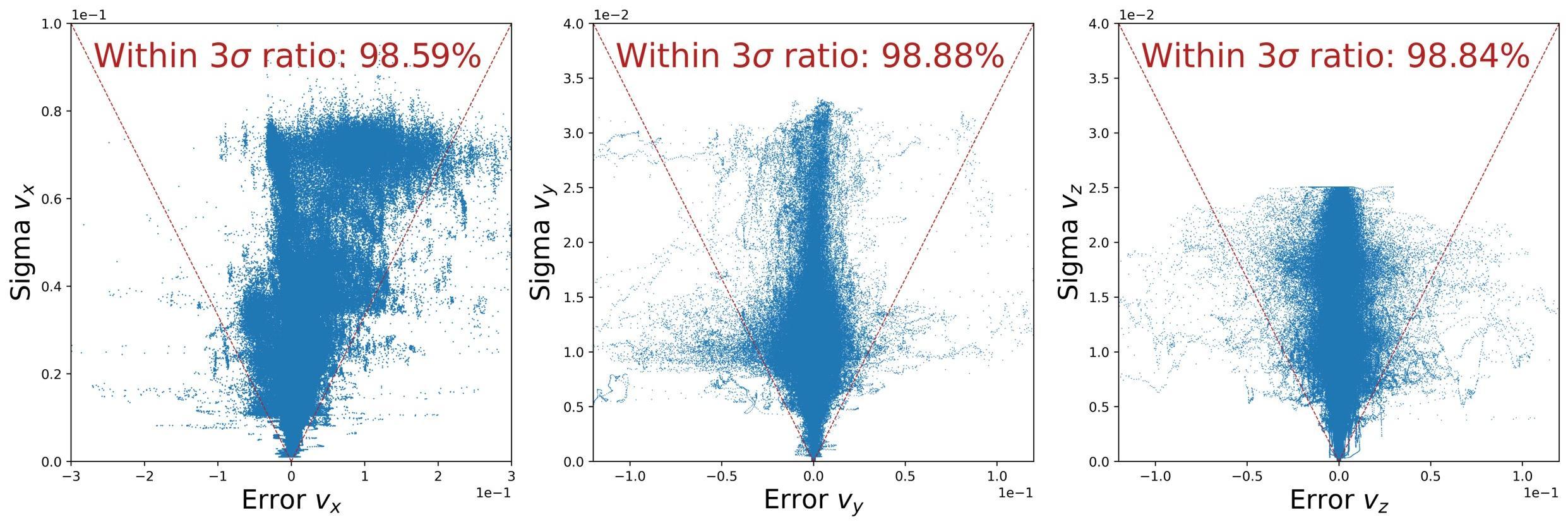}
		\caption{
			Standard deviation $\sigma$ against the errors in ${^{\imuframe}}{\boldsymbol{\widehat{v}}}{_{\imuframe}}_i$ along the x, y, and z axes. 
			Error $v$ represents ${^{\imuframe}}{\boldsymbol{v}}{_{\imuframe}}_i-
			{^{\imuframe}}{\boldsymbol{\widehat{v}}}{_{\imuframe}}_i$. The dashed red line represents the 3$\sigma$ region and the red text indicates the ratio of error points that fall within this 3$\sigma$ region.}
		\label{fig:sigma_v_body}
		\vspace{-0.4cm}
	\end{figure}

	Similar to the consistency analysis performed on the \textit{IMU De-Bias Net}, we have conducted a similar assessment on the \textit{Wheel Encoder Net}. 
	As illustrated in Fig.~\ref{fig:sigma_v_body}, over 98.5\% of the points are situated within the 3$\sigma$ region, indicating the consistency in the outputs generated by the \textit{Wheel Encoder Net}. 
	Due to the skidding, the majority of errors in the x-axis of ${^{\imuframe}}{\boldsymbol{\widehat{v}}}{_{\imuframe}}_i$ are positive. 
	Furthermore, as the vehicle is moving forward, the error and uncertainty along the x-axis are greater than those in the y-axis and z-axis.
	\begin{table}[t]
		\caption{Evaluation of the ${^{\imuframe}}{\boldsymbol{v}}{_{\imuframe}}$ estimation error [$m/s$]. The \textbf{\textcolor{red}{best}} results are marked in red bold.}
		\label{tab:v_body_comparison}
		\vspace{0.0cm}
		\centering
		\resizebox{\linewidth}{!}{
			\begin{tabular}{ccccccc}
				\toprule
				& \multicolumn{3}{c}{\textbf{Raw measurements}} & \multicolumn{3}{c}{\textbf{\textit{Wheel Encoder Net}}} \\
				\toprule
				&${}^{\imuframe}{\boldsymbol{v}}_{\imuframe_x}$ &${}^{\imuframe}{\boldsymbol{v}}_{\imuframe_y}$ &${}^{\imuframe}{\boldsymbol{v}}_{\imuframe_z}$ &${}^{\imuframe}{\boldsymbol{v}}_{\imuframe_x}$ &${}^{\imuframe}{\boldsymbol{v}}_{\imuframe_y}$ &${}^{\imuframe}{\boldsymbol{v}}_{\imuframe_z}$  \\ 
				\toprule
				KAIST urban07  
				& 2.7e-2	& 3.6e-2	& 3.8e-3	& \textbf{\textcolor{red}{7.7e-3}}  & \textbf{\textcolor{red}{3.7e-3}}  & \textbf{\textcolor{red}{3.5e-3}} \\
				\toprule
				KAIST urban09  
				& 2.4e-2	& 2.7e-2	& 3.2e-3	& \textbf{\textcolor{red}{7.5e-3}}  & \textbf{\textcolor{red}{3.4e-3}}  & \textbf{\textcolor{red}{3.1e-3}} \\
				\toprule
				KAIST urban11     
				& 1.2e-1	& 3.8e-2	& \textbf{\textcolor{red}{3.7e-3}}	& \textbf{\textcolor{red}{8.2e-2}}  & \textbf{\textcolor{red}{2.2e-3}}  & 3.9e-3 \\
				\toprule
				KAIST urban13     
				& 1.1e-2	& 1.6e-2	& 7.9e-4	& \textbf{\textcolor{red}{2.1e-3}}  & \textbf{\textcolor{red}{7.2e-4}}  & \textbf{\textcolor{red}{6.7e-4}} \\
				\toprule
				KAIST urban15     
				& 3.2e-2	& 4.1e-2	& 1.8e-3	& \textbf{\textcolor{red}{3.9e-3}}  & \textbf{\textcolor{red}{1.4e-3}}  & \textbf{\textcolor{red}{1.6e-3}} \\
				\toprule
				KAIST urban17     
				& 4.0e-2	& 3.5e-2	& 2.9e-3	& \textbf{\textcolor{red}{2.0e-2}}  & \textbf{\textcolor{red}{4.6e-3}}  & \textbf{\textcolor{red}{2.8e-3}} \\
				\toprule
				NCLT 12/08/20     
				& 9.1e-2	& 5.7e-2	& 7.6e-2	& \textbf{\textcolor{red}{8.0e-2}}  & \textbf{\textcolor{red}{1.6e-2}}  & \textbf{\textcolor{red}{7.5e-2}}  \\
				\toprule
				NCLT 12/12/01     
				& 8.2e-2	& 3.5e-2	& 4.3e-2	& \textbf{\textcolor{red}{6.6e-2}}  & \textbf{\textcolor{red}{1.1e-2}}  & \textbf{\textcolor{red}{4.1e-2}} \\
				\toprule
				NCLT 13/04/05     
				& 8.7e-2	& 3.7e-2	& 4.2e-2	& \textbf{\textcolor{red}{7.3e-2}}  & \textbf{\textcolor{red}{1.0e-2}}  & \textbf{\textcolor{red}{4.0e-2}} \\
				\bottomrule
		\end{tabular}}
		\vspace{-0.7cm}
	\end{table}
	
	\subsubsection{Learned covariance}
	To evaluate the impact of learned covariance, we conduct an evaluation by replacing the learned time-varying covariances from \textit{IMU De-bias Net} and \textit{Wheel Encoder Net} with fixed constant covariances.
		Specifically, following the approach used in the ablation study of TLIO~\cite{tlio2020liu}, we determine the fixed covariances $\widehat{\Cov}_{\rotvel}^2$, $\widehat{\Cov}_{\accvel}^2$ and $\widehat{\Cov}_{\boldsymbol{v}}^2$ by multiplying the standard deviation of the error of the network-corrected IMU and velocity with a fine-tuned factor on the test set. The results in the TABLE.~\ref{tab:fix_cov} demonstrate using the learned covariance can get more accurate pose estimation.
	\begin{table}[t]
		\caption{Evaluation of the effectiveness of using learned covariance from \textit{IMU De-Bias Net} and \textit{Wheel Encoder Net}. The \textbf{\textcolor{red}{best}} results are marked in red bold.}
		\label{tab:fix_cov}
		\vspace{0.0cm}
		\centering
		\scalebox{0.825}{
			
				\begin{tabular}{ccccc}
					\toprule
					& \multicolumn{2}{c}{\textbf{Fixed Covariance}} & \multicolumn{2}{c}{\textbf{Learned Covariance}} \\
					\toprule
					&ATE &ARE &ATE &ARE    \\ 
					\toprule
					KAIST urban07  
					& 10.40	& 1.76	& \textbf{\textcolor{red}{8.10}}	& \textbf{\textcolor{red}{1.47}}    \\
					\toprule
					KAIST urban09  
					& 31.52	& 2.25	& \textbf{\textcolor{red}{22.25}}	& \textbf{\textcolor{red}{2.11}}    \\
					\toprule
					KAIST urban11     
					& 64.54	& 2.61	& \textbf{\textcolor{red}{40.25}}	& \textbf{\textcolor{red}{1.50}}    \\
					\toprule
					KAIST urban13     
					& 15.72	& 2.45	& \textbf{\textcolor{red}{14.58}}	& \textbf{\textcolor{red}{1.94}}    \\
					\toprule
					KAIST urban15     
					& 9.38	& 1.43	& \textbf{\textcolor{red}{7.75}}	& \textbf{\textcolor{red}{1.11}}    \\
					\toprule
					KAIST urban17     
					& 21.57	& 1.09	& \textbf{\textcolor{red}{12.64}}	& \textbf{\textcolor{red}{0.57}}    \\
					\toprule
					NCLT 12/08/20     
					& 18.50	& 3.27	& \textbf{\textcolor{red}{12.35}}	& \textbf{\textcolor{red}{2.91}}    \\
					\toprule
					NCLT 12/12/01     
					& 20.14	& 5.84	& \textbf{\textcolor{red}{9.53}}	& \textbf{\textcolor{red}{2.48}}    \\
					\toprule
					NCLT 13/04/05     
					& 12.97	& 4.30	& \textbf{\textcolor{red}{8.22}}	& \textbf{\textcolor{red}{3.87}}    \\
					\bottomrule
		\end{tabular}}
		\vspace{-0.8cm}
	\end{table}
	\subsubsection{Manifold soft constraint}
	In addition, to further illustrate the effect of imposing ground manifold soft constraint in the EKF framework, we analyze the effects of different manifold models in terms of global continuity and degree, the results of which are listed in the TABLE~\ref{tab:manifold_ablation}.
	It can be evident that the cubic B-spline manifold effectively smoothes the pose of the previously traversed surrounding areas and imposes additional soft constraints on the current pose, and thus the proposed WING system achieves the highest accuracy in pose estimation.
	Under the same order (e.g. a uniform cubic B-spline model), incorporating soft constraints from more traversed areas into a globally continuous manifold model yields superior performance compared to segmented curved manifolds.
	At different orders, the chosen cubic B-spline manifold with $C_2$ continuity surpasses the low-order manifolds which only maintain the $C_0$ or $C_1$ continuity due to the inherent continuity of ground robot accelerations.
	Besides, in some sequences, the system without manifold soft constraints outperforms the system with soft constraints from low-order manifold assumption, which might be due to the fact that the low-order manifolds limit the degrees of freedom for pose estimation. 
	Compared to the segmented manifold, The variation in improvement obtained from using the continuous manifold arises from the extent to which the actual ground adheres to our assumption. 
		Despite the previously traversed surrounding areas generally exhibiting correlation with the current regions, we have to acknowledge that discontinuity parts in the road are unavoidable, often caused by road repairs. 
	
	\begin{table}[t]
		\caption{Ablation study of different manifold models. The \textbf{\textcolor{red}{best}} results are marked in red bold.}
		\label{tab:manifold_ablation}
		\vspace{0.0cm}
		\centering
		\resizebox{\linewidth}{!}{
			\begin{tabular}{ccccccc}
				\toprule
				& ~
				& \textbf{\textit{Segmented}} & \multicolumn{3}{c}{\textbf{\textit{Continuous (B-spline)}}} & \multirow{2}{*}{\textbf{\textit{None}}} \\
				\cmidrule(lr){3-3}\cmidrule(lr){4-6}
				&  &  \textbf{3$^{rd}$} & \textbf{3$^{rd}$ (Ours)} & \textbf{2$^{nd}$} &  \textbf{1$^{st}$} & \\
				\toprule
				KAIST
				& ATE   
				& 8.32  & \textbf{\textcolor{red}{8.10}}   & 9.00  &  9.57  & 8.68   \\
				urban07
				& ARE 
				& 1.70   & \textbf{\textcolor{red}{1.47}}  & 2.19  & 2.20  & 2.10  \\
				\toprule
				KAIST
				& ATE   
				& 25.78 & \textbf{\textcolor{red}{22.25}} & 23.81 & 27.25 & 24.33  \\
				urban09
				& ARE
				& 2.29  & \textbf{\textcolor{red}{2.11}}  & 2.27 & 2.37 & 2.17 \\
				\toprule
				KAIST
				& ATE      
				& 53.27 & \textbf{\textcolor{red}{40.25}} & 49.66  & 64.91 & 44.81  \\
				urban11
				& ARE
				& 1.59  & \textbf{\textcolor{red}{1.50}}   & 1.62 & 2.76 & 1.52 \\
				\toprule
				KAIST
				& ATE      
				& 17.61 & \textbf{\textcolor{red}{14.58}} & 20.33  & 20.68  & 20.22  \\
				urban13
				& ARE 
				& 2.34  & \textbf{\textcolor{red}{1.94}}  & 2.98 & 2.99 & 2.93 \\
				\toprule
				KAIST
				& ATE      
				& 12.77 & \textbf{\textcolor{red}{7.75}}  & 12.31 & 14.77  & 17.79  \\
				urban15
				& ARE
				& 1.41  & \textbf{\textcolor{red}{1.11}}  & 1.40 & 1.92 & 2.66 \\
				\toprule
				KAIST
				& ATE      
				& 29.73 & \textbf{\textcolor{red}{12.64}} & 21.80 & 24.89  & 34.61  \\
				urban17
				& ARE 
				& 1.21  & \textbf{\textcolor{red}{0.57}}  & 1.85  & 1.75 & 2.56	 \\
				\toprule
				NCLT
				& ATE      
				& 22.10 & \textbf{\textcolor{red}{12.35}} & 17.58 & 18.63 & 19.27 \\
				12/08/20
				& ARE 
				& 5.03  & \textbf{\textcolor{red}{2.91}} & 3.53 & 4.28 & 4.48	 \\
				\toprule
				NCLT
				& ATE      
				& 14.36 & \textbf{\textcolor{red}{9.53}} & 13.66 & 14.24 & 14.70 \\
				12/12/01
				& ARE 
				&  3.19 & \textbf{\textcolor{red}{2.48}} & 2.87 & 3.19 & 	3.53 \\
				\toprule
				NCLT
				& ATE      
				& 15.38 & \textbf{\textcolor{red}{8.22}} & 13.73 & 16.12 & 16.10 \\
				13/04/05
				& ARE 
				& 5.55  & \textbf{\textcolor{red}{3.87}} & 4.67 & 5.43 & 	5.37 \\
				\bottomrule
		\end{tabular}}
		\vspace{-0.7cm}
	\end{table}
	\subsection{Runtime analysis}
	Table~\ref{run_time} presents an evaluation of the average processing time for each network module and the entire system on a desktop equipped with an NVIDIA RTX 4090 24 GB GPU and an Intel Core i9-13900K CPU.
		The average processing time for the \textit{IMU De-Bias Net} and \textit{Wheel Encoder Net} corresponds to the inference time of each network, providing network-corrected IMU measurements and velocity, respectively. The average processing time for the space-based sliding-window fusion reflects the overall time required by the entire system for each round of extended Kalman filter (EKF), which includes network processing every 20 rounds and observation updates from manifold soft constraints. Consequently, considering that the IMU inputs are received at a frequency of 100Hz, these results demonstrate the real-time capability of our system.
	\vspace{-0.1cm}
	\section{Conclusion and Future Work}
	\subsection{Conclusion}
	In this work, we propose a novel wheel-inertial odometry system that leverages deep neural networks and soft constraints from ground manifold assumptions within a space-based sliding-window filtering framework. 
	Our system is specifically designed for ground robots, and we utilize the assumption of a dual cubic B-spline manifold to provide globally continuous soft constraints.
	To fully exploit the benefits of both the IMU and wheel encoders, we design deep neural networks to reduce IMU biases and compensate for wheel encoder measurement errors caused by skidding and vertical bouncing. Additionally, the networks implicitly calibrate the IMU and wheel encoder, as well as estimate their time-varying uncertainties for better sensor fusion.
	Experiments show that our proposed algorithm outperforms other open-source methods. Besides, the effectiveness of network-corrected velocity, learned covariance and soft constraints from continuous manifold assumption for pose estimation is shown in the ablation studies.
	\begin{table}[t]
		\vspace{0cm}
		\centering
		\caption{Runtime Analysis}
		\resizebox{\linewidth}{!}{
				\begin{tabular}{cccccccccc}
					\toprule
					& Average Processing Time \\ 
					\toprule
					\textit{IMU De-Bias Net}
					& ${0.6 \ ms}$ \\
					\textit{Wheel Encoder Net}
					& ${0.9 \ ms}$ \\
					\toprule
					Space-based sliding-window fusion 
					& ${0.9 \ ms}$ \\
					\bottomrule
		\end{tabular}}
		\label{run_time}
		\vspace{-0.7cm}
	\end{table}
	\subsection{Future work}
	\label{future-work}
	In the proposed system, neural networks currently employed in the system do not consider the ground manifold. However, an extension could involve tightly coupling neural networks and ground manifold by continuously feeding the estimated ground parameters as input to neural networks. This integration may enhance the fusion process and lead to improved performance.
	Besides, in this work, the covariance of added control points and the size of each knot vector in the B-spline manifold are invariant. In the future, we could use a learning-based method to output the dynamic covariance and time-varying knot vector, making the system more adaptive to more variant scenarios.
	\vspace{-0.5cm}
	\begin{appendices}
		\setstretch{0.85}
		\setlength{\jot}{0.5pt}
		\setcounter{equation}{41}
		\section{Proof of the Eq.~\eqref{thm1}}
		\label{Proof}
		Considering the following kinematic models as utilized in the training process of \textit{De-Bias Net} :
		\begin{equation}
			\label{integral_input}
			\small{
				\begin{aligned}
					{\boldsymbol{v}}_{i,\tau}
					&= \int_{i}^{\tau} 
					{_{\imuframe_t}^{\graframe}}\mathbf{R} 
					({^{\imuframe_t}}\widetilde{\accvel} - 
					{^{\imuframe_t}}\widehat{\bias}_{\accvel} -
					\nb_{\accvel}) dt, \\
					{\boldsymbol{q}}_{i,\tau}
					&= \int_{i}^{\tau}  
					\frac{1}{2} {_{\imuframe_t}^{\imuframe_{i}}} {\boldsymbol{q}} ~ 
					({^{\imuframe_t}}\widetilde{\rotvel} - 
					{^{\imuframe_t}}\widehat{\bias}_{\rotvel}-
					\nb_{\rotvel}) dt,
			\end{aligned}}
		\end{equation}
		where 
		$\tau$ is the time stamp between time $i$ and $i+n$,
		${_{\imuframe_t}^{\graframe}}\mathbf{R}$ is the ground truth rotation, 
		${^{\imuframe_t}}\widehat{\bias}_{\accvel}$ and ${^{\imuframe_t}}\widehat{\bias}_{\rotvel}$ are estimated values of the~\textit{De-Bias Net},
		$\nb_{\accvel}$ and $\nb_{\rotvel}$ are the input noise of the integral system in Eq.~\eqref{integral_input}.
		We discretize the above equations to obtain the nominal state equations:
		\begin{equation}
			\small{
				\begin{aligned}
					{\boldsymbol{v}}_{i,k+1}
					&= {\boldsymbol{v}}_{i,k} + 
					{_{\imuframe_k}^{\graframe}}\mathbf{R} 
					({^{\imuframe_k}}\widetilde{\accvel} - 
					{^{\imuframe_k}}\widehat{\bias}_{\accvel}) \Delta t_k, \\
					{\boldsymbol{q}}_{i,k+1}
					&= {\boldsymbol{q}}_{i,k} ~ 
					{\boldsymbol{q}}
					\{({^{\imuframe_t}}\widetilde{\rotvel} - 
					{^{\imuframe_t}}\widehat{\bias}_{\rotvel})\Delta t_k\} ,
			\end{aligned}}
		\end{equation}
		with the following discrete error state models:
		\begin{equation}
			\label{eq_process}
			\small{
				\begin{aligned}
					&\delta{\boldsymbol{v}}_{i,k+1}
					= \delta{\boldsymbol{v}}_{i,k} + {_{\imuframe_k}^{\graframe}}\mathbf{R} \boldsymbol{\mathbf{v_i}}
					= \delta{\boldsymbol{v}}_{i,k} + \boldsymbol{\mathbf{v_i}}, \\
					&\delta \boldsymbol{\theta}_{i,k+1} =
					{\mathbf{R}}^\top
					\{({^{\imuframe_t}}\widetilde{\rotvel} - 
					{^{\imuframe_t}}\widehat{\bias}_{\rotvel})\Delta t_k\} \delta \boldsymbol{\theta}_{i,k} + \boldsymbol{\mathbf{\theta_i}}, \\
					&\boldsymbol{\mathbf{i}} = 
					\begin{bmatrix}
						\boldsymbol{\mathbf{v_i}}\\ \boldsymbol{\mathbf{\theta_i}}
					\end{bmatrix} \sim \mathcal{N}(0,\mathbf{Q_i}), \quad
					\mathbf{Q_i} = \Delta t^2 
					\begin{bmatrix}
						\Cov_{\accvel}^2 & 
						\mathbf{0}_{3} \\
						\mathbf{0}_{3} & 
						\Cov_{\rotvel}^2 
					\end{bmatrix},
			\end{aligned}}
		\end{equation}
		where $k$ and the $k+1$ are the IMU time stamps between time $i$ and $i+n$, $\boldsymbol{\mathbf{i}}$ is the perturbation impulse vector as in~\cite{sola2017quaternion}.To simplify, we assume that each time interval $\Delta t_k$ is equal to $\Delta t$. Note that, there is no error propagation about the rotation since the ground truth ${_{\imuframe}^{\graframe}}\mathbf{R}$ is provided. 
		We combine the two states in the Eq.~\eqref{eq_process} to yield the following error state equation:
		\begin{equation}
			\small{
				\begin{aligned}
					\delta \boldsymbol{x}_{i,k+1} = 
					\begin{bmatrix}
						\delta{\boldsymbol{v}}_{i,k+1} \\ \delta \boldsymbol{\theta}_{i,k+1}
					\end{bmatrix}
					&= \boldsymbol{f}(\boldsymbol{x}_{i,k},\delta{\boldsymbol{x}}_{i,k},\boldsymbol{u}_{k},\mathbf{i})\\
					&= \mathbf{F_x}_{,k}(\boldsymbol{x}_{i,k},\boldsymbol{u}_{k})\delta{\boldsymbol{x}}_{i,k} + 
					\mathbf{F_i}_{,k}~\mathbf{i},
			\end{aligned}}
		\end{equation}
		whose error state prediction equations are written:
		\begin{equation}
			\small{
				\begin{aligned}
					\delta \widehat{\boldsymbol{x}}_{i,k+1} &= \mathbf{F_x}_{,k}(\boldsymbol{x}_{i,k},\boldsymbol{u}_{k}) \delta \widehat{\boldsymbol{x}}_{i,k},\\
					\mathbf{P}_{i,k+1} &= \mathbf{F_x}_{,k} \mathbf{P}_{i,k} \mathbf{F_x}_{,k}^\top + \mathbf{F_i}_{,k} \mathbf{Q_i} \mathbf{F_i}_{,k}^\top,
			\end{aligned}}
		\end{equation}
		where the Jacobi matrices are:
		\begin{equation}
			\small{
				\begin{aligned}
					\mathbf{F_x}_{,k} &= 
					\begin{bmatrix}
						\mathbf{I}_{3} & \mathbf{0}_{3} \\
						\mathbf{0}_{3} & \mathbf{R} ^\top
						\{({^{\imuframe_k}}\widetilde{\rotvel} - 
						{^{\imuframe_k}}\widehat{\bias}_{\rotvel})\Delta t\}
					\end{bmatrix},  
					\mathbf{F_i} &= 
					\begin{bmatrix}
						\mathbf{I}_{3} & \mathbf{0}_{3} \\
						\mathbf{0}_{3} & \mathbf{I}_{3}
					\end{bmatrix},
			\end{aligned}}
		\end{equation}
		so we can obtain the propagation of the covariance as follows:
		\begin{equation}
			\small{
				\begin{aligned}
					\mathbf{P}_{i,i+1} &= \mathbf{F_x}_{,i+1} \mathbf{P}_{i,i} \mathbf{F_x}_{,i+1}^\top + \mathbf{Q_i},\\
					\mathbf{P}_{i,i+2} &= 
					\mathbf{F_x}_{,i+2} \mathbf{F_x}_{,i+1} \mathbf{P}_{i,i} \mathbf{F_x}_{,i+1}^\top \mathbf{F_x}_{,i+2}^\top \\
					&+ 
					\mathbf{F_x}_{,i+2} \mathbf{P}_{i,i} \mathbf{F_x}_{,i+2}^\top + \mathbf{Q_i},\\
					&~~\smash{\vdots} \\
					\mathbf{P}_{i,i+n} &= 
					\mathbf{F_x}_{,i+n} \cdots \mathbf{F_x}_{,i+1} \mathbf{P}_{i,i} \mathbf{F_x}_{,i+1}^\top \cdots \mathbf{F_x}_{,i+n}^\top \\ &+
					\mathbf{F_x}_{,i+n} \cdots \mathbf{F_x}_{,i+2} \mathbf{Q_i} \mathbf{F_x}_{,i+2}^\top \cdots \mathbf{F_x}_{,i+n}^\top 
					\\ &+ \cdots 
					\mathbf{F_x}_{,i+n} \mathbf{Q_i} \mathbf{F_x}_{,i+n}^\top+ \mathbf{Q_i} ,
			\end{aligned}}
		\end{equation}
		Assuming that the initial covariance matrix is:
		\begin{equation}
			\small{
				\begin{aligned}
					\mathbf{P}_{i,i} &= 
					\begin{bmatrix}
						\mu\mathbf{I}_{3} & \mathbf{0}_{3} \\
						\mathbf{0}_{3} 	  & \nu\mathbf{I}_{3}
					\end{bmatrix} , ~ \mu \textgreater 0, ~ \nu \textgreater 0,
			\end{aligned}}
		\end{equation}
		then we can derive:
		\begin{equation}
			\small{
				\begin{aligned}
					\mathbf{P}_{i,i+n} &= 
					\begin{bmatrix}
						\mu \mathbf{I}_{3}+n \Delta t ^2\Cov_{\accvel}^2 & 
						\mathbf{0}_{3} \\
						\mathbf{0}_{3} & 
						\nu \mathbf{I}_{3}+n \Delta t^2 \Cov_{\rotvel}^2 
					\end{bmatrix}.
			\end{aligned}}
		\end{equation}
		Furthermore, since the initial values of each input window are ground truth in training the~\textit{De-Bias Net} and the initial covariance matrix $\mathbf{P}_{i,i}$  is very small (but cannot be set to 0), then we approximate that
		\begin{equation}
			\small{
				\begin{aligned}
					\mathbf{P}_{i,i+n} &=
					\begin{bmatrix}
						\Cov_{\boldsymbol{v}_{i,i+n}}^2 & \mathbf{0}_{3} \\
						\mathbf{0}_{3} 	  				& \Cov_{\boldsymbol{q}_{i,i+n}}^2
					\end{bmatrix} \\
					&\approx n\Delta t^2
					\begin{bmatrix}
						\Cov_{\accvel}^2  & \mathbf{0}_{3} \\
						\mathbf{0}_{3} 	  & \Cov_{\rotvel}^2
					\end{bmatrix}.
			\end{aligned}}
		\end{equation}
		
		As a result, the estimated covariance matrices in Eq.~\eqref{nll_debias} can be used to approximate the theoretical covariance matrices of the IMU data processed by the \textit{De-Bias Net}:
		\begin{equation}
			\small{
				\begin{aligned}
					\frac{1}{n\Delta t^2}
					\begin{bmatrix}
						\widehat{\Cov}_{\boldsymbol{v}_{i,i+n}}^2 \\
						\widehat{\Cov}_{\boldsymbol{q}_{i,i+n}}^2
					\end{bmatrix} \Longrightarrow
					\begin{bmatrix}
						\widehat{\Cov}_{\accvel}^2 \\
						\widehat{\Cov}_{\rotvel}^2
					\end{bmatrix}.
			\end{aligned}}
		\end{equation}
		\section{Jacobian of Observation model}
		\label{jacob_obs}
		The jacobian of Eq.~\eqref{observation_v} can be described as:
		\begin{equation}
			\small{
				\mathbf{H}_{\boldsymbol{v}} = \begin{aligned}
					&\frac{\partial \boldsymbol{h}_{\boldsymbol{v}}(\mathcal{X})}
					{\partial \mathcal{X}}
					=
					\left[
					\mathbf{0}_{3} \quad
					\mathbf{I}_{3} \quad
					\mathbf{0}_{3\times (3+6n+16)}
					\right],\\
			\end{aligned}}
			\label{jacob_obs1}
		\end{equation}
		
		The jacobian of the Eq.~\eqref{observation_static} with respect to the state $\mathcal{X}$ can be described as:
		\begin{equation}
			\small{
				\mathbf{H}_{\boldsymbol{\mathcal{M}_1}} =
				\begin{aligned}
					&\frac{\partial \boldsymbol{h}_{u,v}(\mathcal{X})}
					{\partial \mathcal{X}} = 
					\begin{bmatrix}
						\mathbf{0}_{3n_{u,v}\times 9} \quad
						\mathbf{0}_{3n_{u,v}\times 6n_{u,v}} \quad
						\frac{\partial \boldsymbol{h}_{u,v}(\mathcal{X})}
						{\partial \boldsymbol{c}} 
					\end{bmatrix}.
			\end{aligned}}
			\label{jacob_obs2_overall}
		\end{equation}
		Specifically,
		\begin{subequations}
			\small{
				\begin{align}
					&\frac{\partial \boldsymbol{h}_{u,v}(\mathcal{X})}
					{\partial \boldsymbol{c}} =
					\begin{bmatrix}
						\frac{\partial \boldsymbol{h}_{u,v,\boldsymbol{\xi}_1}(\mathcal{X})}
						{\partial \boldsymbol{c}} \\
						\vdots \\
						\frac{\partial \boldsymbol{h}_{u,v,\boldsymbol{\xi}_{n_{u,v}}}(\mathcal{X})}
						{\partial \boldsymbol{c}}
					\end{bmatrix}_{3n_{u,v}\times 16},\\
					&\frac{\partial \boldsymbol{h}_{u,v,\boldsymbol{\xi}}(\mathcal{X})}
					{\partial \boldsymbol{c}} =
					\begin{bmatrix}
						\boldsymbol{y}_w \mathbf{K}_y \mathbf{B} \otimes 
						\boldsymbol{x}_w \mathbf{K}_x \mathbf{B} \mathbf{M}_S(u,v) \\
						\boldsymbol{\partial y}_w \mathbf{K}_y \mathbf{B} \otimes 
						\boldsymbol{x}_w \mathbf{K}_x \mathbf{B} \mathbf{M}_S(u,v) \\
						\boldsymbol{y}_w \mathbf{K}_y \mathbf{B} \otimes 
						\boldsymbol{\partial x}_w \mathbf{K}_x \mathbf{B} \mathbf{M}_S(u,v)
					\end{bmatrix}_{3\times 16}.
			\end{align}}
			\label{jacob_obs2}
		\end{subequations}
		
		The jacobian of the Eq.~\eqref{observation_active} with respect to the state $\mathcal{X}$ can be described as:
		\begin{equation}
			\small{
				\begin{aligned}
					\mathbf{H}_{\boldsymbol{\mathcal{M}_2}} = &\frac{\partial \boldsymbol{h}_{g_x,g_y}(\mathcal{X})}
					{\partial \mathcal{X}} = 
					\begin{bmatrix}
						\mathbf{0}_{3n_{g_x,g_y}\times 9} \quad
						\frac{\partial \boldsymbol{h}_{g_x,g_y}(\mathcal{X})}
						{\partial \mathcal{X}_S} \quad
						\frac{\partial \boldsymbol{h}_{g_x,g_y}(\mathcal{X})}
						{\partial \boldsymbol{c}} 
					\end{bmatrix}.
			\end{aligned}}
			\label{jacob_obs3_overall}
		\end{equation}
		Specifically,
		\begin{subequations}
			\small{
				\begin{align}
					&\frac{\partial \boldsymbol{h}_{g_x,g_y}(\mathcal{X})}
					{\partial \mathcal{X}_S} = \nonumber \\
					&\setlength{\arraycolsep}{0.7pt}
					\begin{bmatrix}
						\frac{\partial \boldsymbol{h}_{g_x,g_y,\boldsymbol{\xi}_1}(\mathcal{X})}
						{\partial \boldsymbol{\xi}_1} & \mathbf{0}_{3\times 6} & \cdots & \mathbf{0}_{3\times 6} \\
						\mathbf{0}_{3\times 6} & 
						\frac{\partial \boldsymbol{h}_{g_x,g_y,\boldsymbol{\xi}_2}(\mathcal{X})}
						{\partial \boldsymbol{\xi}_2} & \cdots & \mathbf{0}_{3\times 6} \\
						\vdots & \vdots & \ddots & \vdots \\
						\mathbf{0}_{3\times 6} & \mathbf{0}_{3\times 6} & \cdots &
						\frac{\partial \boldsymbol{h}_{g_x,g_y,\boldsymbol{\xi}_{n_{g_x,g_y}}}(\mathcal{X})}
						{\partial \boldsymbol{\xi}_{n_{g_x,g_y}}}
					\end{bmatrix},\\
					&\frac{\partial \boldsymbol{h}_{g_x,g_y}(\mathcal{X})}
					{\partial \boldsymbol{c}} =
					\begin{bmatrix}
						\frac{\partial \boldsymbol{h}_{g_x,g_y,\boldsymbol{\xi}_1}(\mathcal{X})}
						{\partial \boldsymbol{c}} \\
						\vdots \\
						\frac{\partial \boldsymbol{h}_{g_x,g_y,\boldsymbol{\xi}_{n_{u,v}}}(\mathcal{X})}
						{\partial \boldsymbol{c}}
					\end{bmatrix}_{3n_{g_x,g_y}\times 16},\\
					&\frac{\partial \boldsymbol{h}_{g_x,g_y,\boldsymbol{\xi}}(\mathcal{X})}
					{\partial \boldsymbol{c}} =
					\begin{bmatrix}
						\boldsymbol{y}_w \mathbf{K}_y \mathbf{B} \otimes 
						\boldsymbol{x}_w \mathbf{K}_x \mathbf{B} \\
						\boldsymbol{\partial y}_w \mathbf{K}_y \mathbf{B} \otimes 
						\boldsymbol{x}_w \mathbf{K}_x \mathbf{B} \\
						\boldsymbol{y}_w \mathbf{K}_y \mathbf{B} \otimes 
						\boldsymbol{\partial x}_w \mathbf{K}_x \mathbf{B} 
					\end{bmatrix}_{3\times 16},\\
					&\frac{\partial \boldsymbol{h}_{g_x,g_y,\boldsymbol{\xi}}(\mathcal{X})}
					{\partial \boldsymbol{\xi}} =
					\setlength{\arraycolsep}{0.8pt}
					\begin{bmatrix}
						\frac{\partial \boldsymbol{h}_{g_x,g_y,\boldsymbol{\xi}}(\mathcal{X})}
						{\partial \boldsymbol{p}} & 
						\frac{\partial \boldsymbol{h}_{g_x,g_y,\boldsymbol{\xi}}(\mathcal{X})}
						{\partial \psi} & 
						\frac{\partial \boldsymbol{h}_{g_x,g_y,\boldsymbol{\xi}}(\mathcal{X})}
						{\partial \boldsymbol{s}} 
					\end{bmatrix}_{3\times 6},\\ 
					&\frac{\partial \boldsymbol{h}_{g_x,g_y,\boldsymbol{\xi}}(\mathcal{X})}
					{\partial \boldsymbol{p}} = 
					\begin{bmatrix}
						\boldsymbol{\partial x}_w \mathbf{G} \boldsymbol{y}_w^\top &
						\boldsymbol{x}_w \mathbf{G} \boldsymbol{\partial y}_w^\top &
						-1 \\
						\boldsymbol{\partial x}_w \mathbf{G} \boldsymbol{\partial y}_w^\top &
						\boldsymbol{x}_w \mathbf{G} \boldsymbol{\partial^2 y}_w^\top &
						0 \\
						\boldsymbol{\partial^2 x}_w \mathbf{G} \boldsymbol{y}_w^\top &
						\boldsymbol{\partial x}_w \mathbf{G} \boldsymbol{\partial y}_w^\top &
						0
					\end{bmatrix}, \\
					&\frac{\partial \boldsymbol{h}_{g_x,g_y,\boldsymbol{\xi}}(\mathcal{X})}
					{\partial \psi} =
					\begin{bmatrix}
						\boldsymbol{\partial x}_w \mathbf{G} \boldsymbol{y}_w^\top &
						\boldsymbol{x}_w \mathbf{G} \boldsymbol{\partial y}_w^\top &
						-1 \\
						\boldsymbol{\partial x}_w \mathbf{G} \boldsymbol{\partial y}_w^\top &
						\boldsymbol{x}_w \mathbf{G} \boldsymbol{\partial^2 y}_w^\top &
						0 \\
						\boldsymbol{\partial^2 x}_w \mathbf{G} \boldsymbol{y}_w^\top &
						\boldsymbol{\partial x}_w \mathbf{G} \boldsymbol{\partial y}_w^\top &
						0
					\end{bmatrix} 
					\frac{\partial \mathbf{R} \boldsymbol{t}}
					{\partial \psi} \nonumber \\ 
					&+
					2\begin{bmatrix}
						0\\
						\frac{s_1\cos \psi-s_2\sin \psi}{1-s_1^2-s_2^2} \\
						\frac{-s_1\sin \psi-s_2\cos \psi}{1-s_1^2-s_2^2}
					\end{bmatrix},\\
					&\frac{\partial \boldsymbol{h}_{g_x,g_y,\boldsymbol{\xi}}(\mathcal{X})}
					{\partial \boldsymbol{s}} = 
					\begin{bmatrix}
						\boldsymbol{\partial x}_w \mathbf{G} \boldsymbol{y}_w^\top &
						\boldsymbol{x}_w \mathbf{G} \boldsymbol{\partial y}_w^\top &
						-1 \\
						\boldsymbol{\partial x}_w \mathbf{G} \boldsymbol{\partial y}_w^\top &
						\boldsymbol{x}_w \mathbf{G} \boldsymbol{\partial^2 y}_w^\top &
						0 \\
						\boldsymbol{\partial^2 x}_w \mathbf{G} \boldsymbol{y}_w^\top &
						\boldsymbol{\partial x}_w \mathbf{G} \boldsymbol{\partial y}_w^\top &
						0
					\end{bmatrix} 
					\frac{\partial \mathbf{R} \boldsymbol{t}}
					{\partial \boldsymbol{s}} \nonumber \\ 
					&+2\begin{bmatrix}
						0 & 0 \\
						\frac{(1+s_1^2-s_2^2)\sin \psi + 2s_1s_2\cos \psi}{(1-s_1^2-s_2^2)^2} &
						\frac{(1-s_1^2+s_2^2)\cos \psi + 2s_1s_2\sin \psi}{(1-s_1^2-s_2^2)^2} \\
						\frac{(1+s_1^2-s_2^2)\sin \psi - 2s_1s_2\cos \psi}{(1-s_1^2-s_2^2)^2} &
						\frac{(-1+s_1^2-s_2^2)\cos \psi + 2s_1s_2\sin \psi}{(1-s_1^2-s_2^2)^2} 
					\end{bmatrix}, \\
					&\frac{\partial \mathbf{R} \boldsymbol{t}}
					{\partial \psi} =
					\begin{bmatrix}
						-\sin \psi & \cos \psi & 0\\
						\cos \psi & -\sin \psi & 0\\
						0 & 0 & 0 
					\end{bmatrix}\mathbf{R}_{\phi} \boldsymbol{t}, \\
					&\frac{\partial \mathbf{R} \boldsymbol{t}}
					{\partial s_1} =\mathbf{R}_\psi
					\setlength{\arraycolsep}{0.8pt}
					\begin{bmatrix}
						\frac{-4s_1(1+s_2^2}{1+{s_{1}^2}+{s_{2}^2}}  & \frac{-2s_2(1-s_1^2+s_2^2)}{1+{s_{1}^2}+{s_{2}^2}} & \frac{2(1-s_1^2+s_2^2)}{1+{s_{1}^2}+{s_{2}^2}} \\
						\frac{-2s_2(1-s_1^2+s_2^2)}{1+{s_{1}^2}+{s_{2}^2}} & \frac{4s_1s_2^2}{1+{s_{1}^2}+{s_{2}^2}} 			& \frac{-4s_1s_2}{1+{s_{1}^2}+{s_{2}^2}} \\
						\frac{-2(1-s_1^2+s_2^2)}{1+{s_{1}^2}+{s_{2}^2}} 	 &  \frac{4s_1s_2}{1+{s_{1}^2}+{s_{2}^2}} 			& \frac{-4s_1}{1+{s_{1}^2}+{s_{2}^2}}
					\end{bmatrix}\boldsymbol{t}, \\
					&\frac{\partial \mathbf{R} \boldsymbol{t}}
					{\partial s_2} = 
					\mathbf{R}_\psi
					\setlength{\arraycolsep}{0.8pt}
					\begin{bmatrix}
						\frac{4s_1^2s_2}{1+{s_{1}^2}+{s_{2}^2}} 			 & \frac{-2s_2(1+s_1^2-s_2^2)}{1+{s_{1}^2}+{s_{2}^2}} & \frac{-4s_1s_2}{1+{s_{1}^2}+{s_{2}^2}} \\
						\frac{-2s_2(1+s_1^2-s_2^2)}{1+{s_{1}^2}+{s_{2}^2}} & \frac{-4s_2(1+s_1^2)}{1+{s_{1}^2}+{s_{2}^2}} 		& \frac{2(1+s_1^2-s_2^2)}{1+{s_{1}^2}+{s_{2}^2}} \\
						\frac{4s_1s_2}{1+{s_{1}^2}+{s_{2}^2}} 			 &  \frac{-2(1+s_1^2-s_2^2)}{1+{s_{1}^2}+{s_{2}^2}} 	& \frac{-4s_2}{1+{s_{1}^2}+{s_{2}^2}}
					\end{bmatrix}\boldsymbol{t},
				\end{align}
			}
			\label{jacob_obs3}
		\end{subequations}
		where 
		$\{\mathbf{R},
		\mathbf{R}_\psi,
		\mathbf{R}_\phi,
		\boldsymbol{t}\}$ are the abbreviations of 
		$\{{}_{\imuframe}^{\graframe} \mathbf{R},
		{}_{\imuframe}^{\graframe} \mathbf{R}_\psi,
		{}_{\imuframe}^{\graframe} \mathbf{R}_\phi,
		{}^{\imuframe} \boldsymbol{t}_{w}\}$, $\mathbf{G} = \mathbf{K}_x \mathbf{B} \mathbf{C}
		\mathbf{B}^\top \mathbf{K}_y^\top$.
		\section{Jacobian of Process model}	
		\label{jacob_fx}
		The jacobian of Eq.~\eqref{discrete_process_model} can be described as:
		\begin{subequations} 
			\label{jacob_fx_overall}
			\small{
				\setlength{\arraycolsep}{1.0pt}
				\begin{align}
					\mathbf{F}_{\imuframe} &=
					\begin{bmatrix} 
						\mathbf{I}_{3}	
						&{{}_{\imuframe_{k}}^{\graframe}}\mathbf{R} \Delta t_{k}
						&\frac{\partial {{}^{\graframe}} \boldsymbol{p}_{\imuframe_{k+1}}}
						{\partial {{}^{\graframe}_{\imuframe_{k}}} \psi}
						&\frac{\partial {{}^{\graframe}} \boldsymbol{p}_{\imuframe_{k+1}}}
						{\partial {{}^{\graframe}_{\imuframe_{k}}} \boldsymbol{s}}\\
						\mathbf{0}_{3}	
						&\mathbf{I}_{3}	- \left\lfloor \boldsymbol{\omega}_k \right\rfloor _\times \Delta t_{k}
						&\mathbf{0}_{3\times 1}	
						&\frac{\partial {{}^{\imuframe_{k+1}}} \boldsymbol{v}}
						{\partial {{}^{\graframe}_{\imuframe_{k}}} \boldsymbol{s}}\\[0.5em]
						\mathbf{0}_{1\times 3}	
						&\mathbf{0}_{1\times 3}	
						& 1
						&\frac{\partial {{}^{\graframe}_{\imuframe_{k+1}}} \psi}
						{\partial {{}^{\graframe}_{\imuframe_{k}}} \boldsymbol{s}}\\[0.5em]
						\mathbf{0}_{2\times 3}	
						&\mathbf{0}_{2\times 3}
						&\mathbf{0}_{2\times 1}
						&\frac{\partial {{}^{\graframe}_{\imuframe_{k+1}}} \boldsymbol{s}}
						{\partial {{}^{\graframe}_{\imuframe_{k}}} \boldsymbol{s}}\\[0.5em]
					\end{bmatrix}, \\
					\mathbf{F}_{\boldsymbol{n}} &=
					\begin{bmatrix} 
						\mathbf{0}_{3}	  	
						&\mathbf{0}_{3} \\
						-\mathbf{I}_{3} \Delta t_{k}
						&\quad-\left\lfloor {{}^{\imuframe}} \boldsymbol{v} 
						\right\rfloor _\times  \Delta t_{k} \\
						\mathbf{0}_{1\times 3}	  	
						&\frac{\partial {{}^{\graframe}_{\imuframe_{k+1}}} \psi}
						{\partial \nb_{\boldsymbol{\omega}}} \\
						\mathbf{0}_{2\times 3}	  	
						&\frac{\partial {{}^{\graframe}_{\imuframe_{k+1}}} \boldsymbol{s}}
						{\partial \nb_{\boldsymbol{\omega}}}
					\end{bmatrix}.  
			\end{align}}
		\end{subequations} 
		Specifically,
		\begin{subequations} 
			\small{
				\begin{align}
					&\frac{\partial {{}^{\graframe}} \boldsymbol{p}_{\imuframe_{k+1}}}
					{\partial {{}^{\graframe}_{\imuframe_{k}}} \psi}
					= 
					\begin{bmatrix}
						-\sin \psi &
						-\cos \psi &
						0 \\
						\cos \psi &
						-\sin \psi &
						0 \\
						0 & 0 & 0
					\end{bmatrix}
					\mathbf{R}_{\phi} {{}^{\imuframe}} \boldsymbol{v} \Delta t_{k},\\
					&\frac{\partial {{}^{\graframe}} \boldsymbol{p}_{\imuframe_{k+1}}}
					{\partial {{}^{\graframe}_{\imuframe_{k}}} {s_{1}}} 
					\!=\!
					\setlength{\arraycolsep}{0.7pt}
					\small{
						\mathbf{R}_\psi \!
						\begin{bmatrix}
							\frac{-4s_1(1+s_2^2)}{1+{s_{1}^2}+{s_{2}^2}}  & \frac{-2s_2(1-s_1^2+s_2^2)}{1+{s_{1}^2}+{s_{2}^2}} & \frac{2(1-s_1^2+s_2^2)}{1+{s_{1}^2}+{s_{2}^2}} \\
							\frac{-2s_2(1-s_1^2+s_2^2)}{1+{s_{1}^2}+{s_{2}^2}} & \frac{4s_1s_2^2}{1+{s_{1}^2}+{s_{2}^2}} 			& \frac{-4s_1s_2}{1+{s_{1}^2}+{s_{2}^2}} \\ 
							\frac{-2(1-s_1^2+s_2^2)}{1+{s_{1}^2}+{s_{2}^2}} 	 &  \frac{4s_1s_2}{1+{s_{1}^2}+{s_{2}^2}} 			& \frac{-4s_1}{1+{s_{1}^2}+{s_{2}^2}}
						\end{bmatrix}
					}\small
					\!{{}^{\imuframe}} \boldsymbol{v} \Delta t_{k}, \\
					&\frac{\partial {{}^{\graframe}} \boldsymbol{p}_{\imuframe_{k+1}}}
					{\partial {{}^{\graframe}_{\imuframe_{k}}} {s_{2}}}
					\!=\!
					\small{ 
						\mathbf{R}_\psi \!
						\setlength{\arraycolsep}{0.7pt}
						\begin{bmatrix}
							\frac{4s_1^2s_2}{1+{s_{1}^2}+{s_{2}^2}} 			 & \frac{-2s_2(1+s_1^2-s_2^2)}{1+{s_{1}^2}+{s_{2}^2}} & \frac{-4s_1s_2}{1+{s_{1}^2}+{s_{2}^2}} \\
							\frac{-2s_2(1+s_1^2-s_2^2)}{1+{s_{1}^2}+{s_{2}^2}} & \frac{-4s_2(1+s_1^2)}{1+{s_{1}^2}+{s_{2}^2}} 		& \frac{2(1+s_1^2-s_2^2)}{1+{s_{1}^2}+{s_{2}^2}} \\
							\frac{4s_1s_2}{1+{s_{1}^2}+{s_{2}^2}} 			 &  \frac{-2(1+s_1^2-s_2^2)}{1+{s_{1}^2}+{s_{2}^2}} 	& \frac{-4s_2}{1+{s_{1}^2}+{s_{2}^2}}
						\end{bmatrix}
					}\small
					\!{{}^{\imuframe}} \boldsymbol{v} \Delta t_{k},\\
					\!&\frac{\partial {{}^{\imuframe_{k+1}}} \boldsymbol{v}}
					{\partial {{}^{\graframe}_{\imuframe_{k}}} \boldsymbol{s}}
					= 
					\small{
						-g
						\begin{bmatrix}
							\frac{-2(1-s_1^2+s_2^2)}{1+{s_{1}^2}+{s_{2}^2}} & \frac{4s_1s_2}{1+{s_{1}^2}+{s_{2}^2}}\\
							\frac{4s_1s_2}{1+{s_{1}^2}+{s_{2}^2}} 		  & \frac{-2(1+s_1^2-s_2^2)}{1+{s_{1}^2}+{s_{2}^2}}\\ 
							\frac{-4s_1}{1+{s_{1}^2}+{s_{2}^2}} 			  & \frac{-4s_2}{1+{s_{1}^2}+{s_{2}^2}}
						\end{bmatrix}\Delta t_{k}
					}\small,\\
					&\frac{\partial {{}^{\graframe}_{\imuframe_{k+1}}} \psi}
					{\partial {{}^{\graframe}_{\imuframe_{k}}} \boldsymbol{s}}
					= - 
					\begin{bmatrix}
						\omega_1 & \omega_2
					\end{bmatrix} \Delta t_{k}, \\
					&\frac{\partial {{}^{\graframe}_{\imuframe_{k+1}}} \boldsymbol{s}}
					{\partial {{}^{\graframe}_{\imuframe_{k}}} \boldsymbol{s}}
					=
					\small{
						\mathbf{I}_{2}+
						\begin{bmatrix}
							\omega_2s_1-\omega_1s_2 		&-\omega_1s_1-\omega_2s_2+\omega_3 \\
							\omega_1s1+\omega_2s_2-\omega_3 &\omega_2s_1-\omega_1s_2
						\end{bmatrix}\Delta t_{k}
					}\small, \\
					&\frac{\partial {{}^{\graframe}_{\imuframe_{k+1}}} \psi}
					{\partial \nb_{\boldsymbol{\omega}}}
					= \begin{bmatrix}
						s_1 & s_2 & -1
					\end{bmatrix}\Delta t_{k}, \\
					&\frac{\partial {{}^{\graframe}_{\imuframe_{k+1}}} \boldsymbol{s}}
					{\partial \nb_{\boldsymbol{\omega}}}
					= \begin{bmatrix}
						s_1s_2 & \frac{1}{2}(-1-s_1^2+s_2^2) & -s_2 \\
						\frac{1}{2}(1-s_1^2+s_2^2) & -s_1s_2 &  s_1 \\
					\end{bmatrix}\Delta t_{k}, 
				\end{align}
			}\small
			\label{jacob_process}
		\end{subequations} 
		where 
		$\{
		{{}^{\imuframe}} \boldsymbol{v} ,
		\psi ,
		\phi , 
		s_1 , 
		s_2 ,
		\omega_1,
		\omega_2,
		\omega_3
		\}$ 
		are the abbreviations of
		$\{
		{{}^{\imuframe_{k}}} \boldsymbol{v} ,
		{{}^{\graframe}_{\imuframe_{k}}} \psi , 
		{{}^{\graframe}_{\imuframe_{k}}} \phi , 
		{{}^{\graframe}_{\imuframe_{k}}} s_1 ,
		{{}^{\graframe}_{\imuframe_{k}}} s_2 ,
		{{}^{\imuframe_{k}}} \omega_1,
		{{}^{\imuframe_{k}}} \omega_2,
		{{}^{\imuframe_{k}}} \omega_3
		\}$ 
		and $\left\lfloor \boldsymbol{a} \right\rfloor _\times$ is the skew symmetric matrix of any vector $\boldsymbol{a}$.

		\section{The derivation of Equation~\eqref{manifold_constraint_simple}}
		\label{derivation_obs}
		Based on the Eq.~\eqref{att_mat}, we can get:
		\begin{equation}
			\begin{aligned}
				{}_{w}^{\graframe}\mathbf{R} \cdot {\boldsymbol{e}_3} &= 
				\begin{bmatrix}
					\cos\psi  & -\sin\psi & 0 \\ 
					\sin \psi & \cos \psi & 0 \\ 
					0 & 0 & 1 
				\end{bmatrix}
				\begin{bmatrix}
					\frac{2 s_{1}}{1+s_{1}^2+s_{2}^2} \\
					\frac{2 s_{2}}{1+s_{1}^2+s_{2}^2} \\ 
					\frac{1-s_{1}^2-s_{2}^2}{1+s_{1}^2+s_{2}^2}
				\end{bmatrix} \\
				&=		
				\begin{bmatrix}
					\cos\psi \frac{2 s_{1}}{1+s_{1}^2+s_{2}^2} - \sin\psi \frac{2 s_{2}}{1+s_{1}^2+s_{2}^2} \\
					\sin\psi \frac{2 s_{1}}{1+s_{1}^2+s_{2}^2} + \cos\psi \frac{2 s_{2}}{1+s_{1}^2+s_{2}^2} \\ 
					\frac{1-s_{1}^2-s_{2}^2}{1+s_{1}^2+s_{2}^2}
				\end{bmatrix}
			\end{aligned}
		\end{equation}
		Based on Eq.~\eqref{manifold_simple}, we can get:
		\begin{equation}
			\begin{aligned}
				\nabla \mathcal{M} \left({}^{\graframe}\boldsymbol{p}_w \right) &= [\frac{\partial {}^{\graframe}\boldsymbol{p}_w}{\partial \boldsymbol{x}}, \frac{\partial {}^{\graframe}\boldsymbol{p}_w}{\partial \boldsymbol{y}}, -1]^\top \\
				&= 	
				\begin{bmatrix}
					\partial \boldsymbol{x}_w \mathbf{K}_x \mathbf{B} \mathbf{C} \mathbf{B}^\top \mathbf{K}_y^\top \boldsymbol{y}_w^\top \\
					\boldsymbol{x}_w \mathbf{K}_x \mathbf{B} \mathbf{C} \mathbf{B}^\top \mathbf{K}_y^\top \boldsymbol{\partial y}_w^\top \\
					-1
				\end{bmatrix}
			\end{aligned}		
		\end{equation}
		Then, Eq.~\eqref{manifold_constraint} can be written as:
		\begin{equation}
			\begin{aligned}
				&\mathcal{M} \left({}^{\graframe}\boldsymbol{p}_w\right) = 
				\boldsymbol{x}_w \mathbf{K}_x \mathbf{B} \mathbf{C} \mathbf{B}^\top \mathbf{K}_y^\top \boldsymbol{y}^\top - z_w = 0,\\
				&({}_{w}^{\graframe}\mathbf{R} \cdot {\boldsymbol{e}_3}) \times
				\nabla \mathcal{M} \left({}^{\graframe}\boldsymbol{p}_w \right) = \\
				&\begin{bmatrix}
					\boldsymbol{x}_w \mathbf{K}_x \mathbf{B} \mathbf{C} \mathbf{B}^\top \mathbf{K}_y^\top \boldsymbol{\partial y}_w^\top + 2 \frac{s_1 \sin(\psi) + s_2 \cos(\psi)}{1-s_1^2-s_2^2} \\
					\boldsymbol{\partial x}_w \mathbf{K}_x \mathbf{B} \mathbf{C} \mathbf{B}^\top \mathbf{K}_y^\top \boldsymbol{y}_w^\top + 2 \frac{s_1 \cos(\psi) - s_2 \sin(\psi)}{1-s_1^2-s_2^2} \\ 
					/
				\end{bmatrix}&= \mathbf{0},
			\end{aligned}
		\end{equation}
		Since the rank of $({}_{w}^{\graframe}\mathbf{R} \cdot {\boldsymbol{e}_3}) \times
		\nabla \mathcal{M} \left({}^{\graframe}\boldsymbol{p}_w \right)$ is $2$, only the first two rows are taken into account. The "$/$" symbol indicates that the remaining part of the equation is not considered.
		Therefore, we can get Eq.~\eqref{manifold_constraint_simple}:
		\begin{equation}
			\label{manifold_constraint_simple}
			\begin{aligned}
				\boldsymbol{x}_w \mathbf{K}_x \mathbf{B} \mathbf{C} \mathbf{B}^\top \mathbf{K}_y^\top \boldsymbol{y}^\top &- z_w = 0, \\ 
				\boldsymbol{x}_w \mathbf{K}_x \mathbf{B} \mathbf{C} \mathbf{B}^\top \mathbf{K}_y^\top \boldsymbol{\partial y}_w^\top &+ 2 \frac{s_1 \sin(\psi) + s_2 \cos(\psi)}{1-s_1^2-s_2^2} = 0,\\ 
				\boldsymbol{\partial x}_w \mathbf{K}_x \mathbf{B} \mathbf{C} \mathbf{B}^\top \mathbf{K}_y^\top \boldsymbol{y}_w^\top &+ 2 \frac{s_1 \cos(\psi) - s_2 \sin(\psi)}{1-s_1^2-s_2^2} = 0,
			\end{aligned}
		\end{equation}

	\end{appendices}
	\bibliography{paper}
	\begin{IEEEbiography}[{\includegraphics[width=1in,height=1.25in,clip,keepaspectratio]{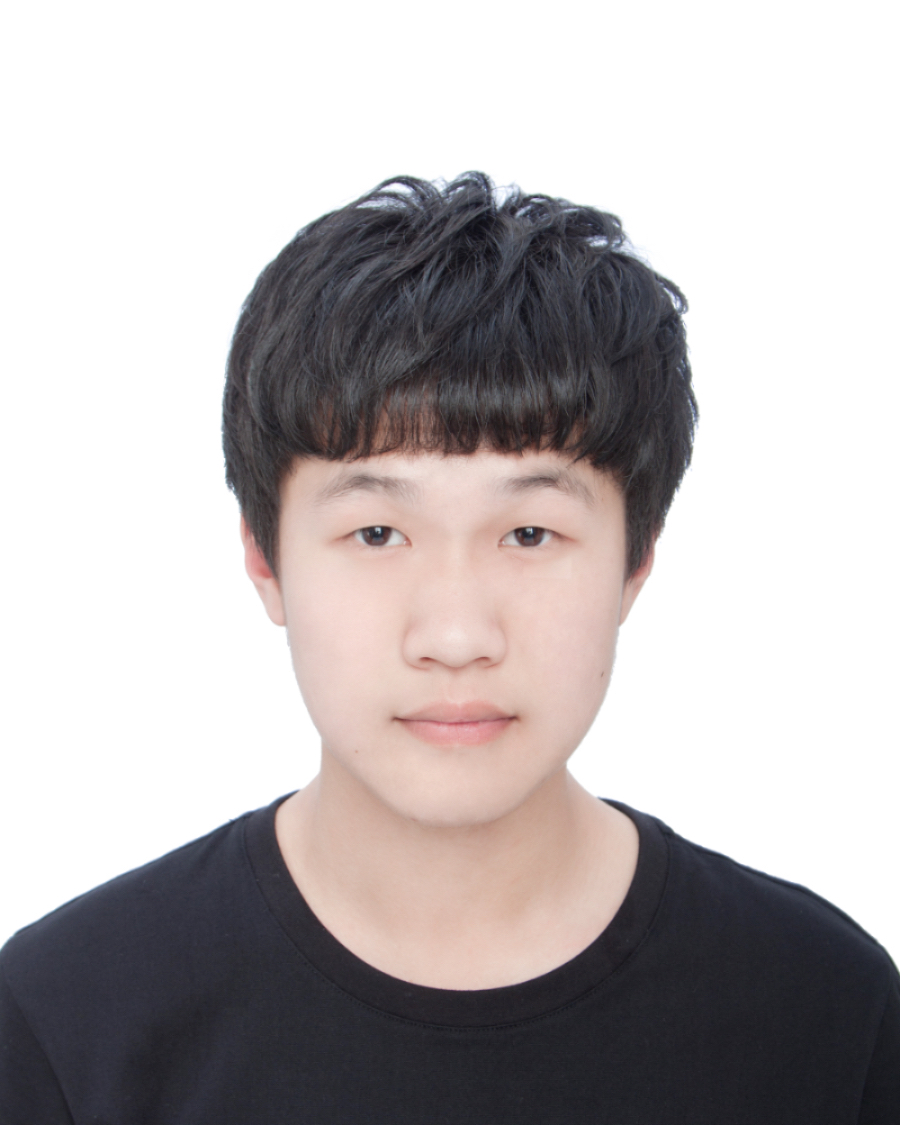}}]{Chenxing Jiang} received the B.Eng. degree in Automation from the Zhejiang University, China, in 2022. 
		
		He then joined HKUST Aerial Robotics Group, The Hong Kong University of Science and Technology, under the supervision of Prof. Shaojie Shen. His research interests include 3D reconstruction, SLAM and robot learning.
	\end{IEEEbiography}
	\begin{IEEEbiography}[{\includegraphics[width=1in,height=1.25in,clip,keepaspectratio]{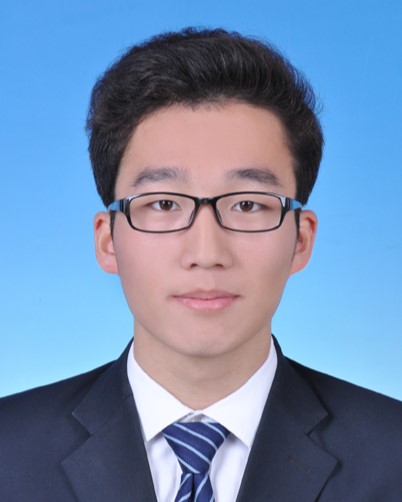}}]{Kunyi Zhang} received the Eng.D. degree in electronic and information engineering at the College of Control Science and Engineering from Zhejiang University, China, in 2023.
		
		He is currently a post-doctoral researcher at the School of Mechanical Engineering, Zhejiang University. 
		His research interests include state estimation, sensor fusion, localization and mapping, and autonomous navigation.
	\end{IEEEbiography}
	
	\begin{IEEEbiography}[{\includegraphics[width=1in,height=1.25in,clip,keepaspectratio]{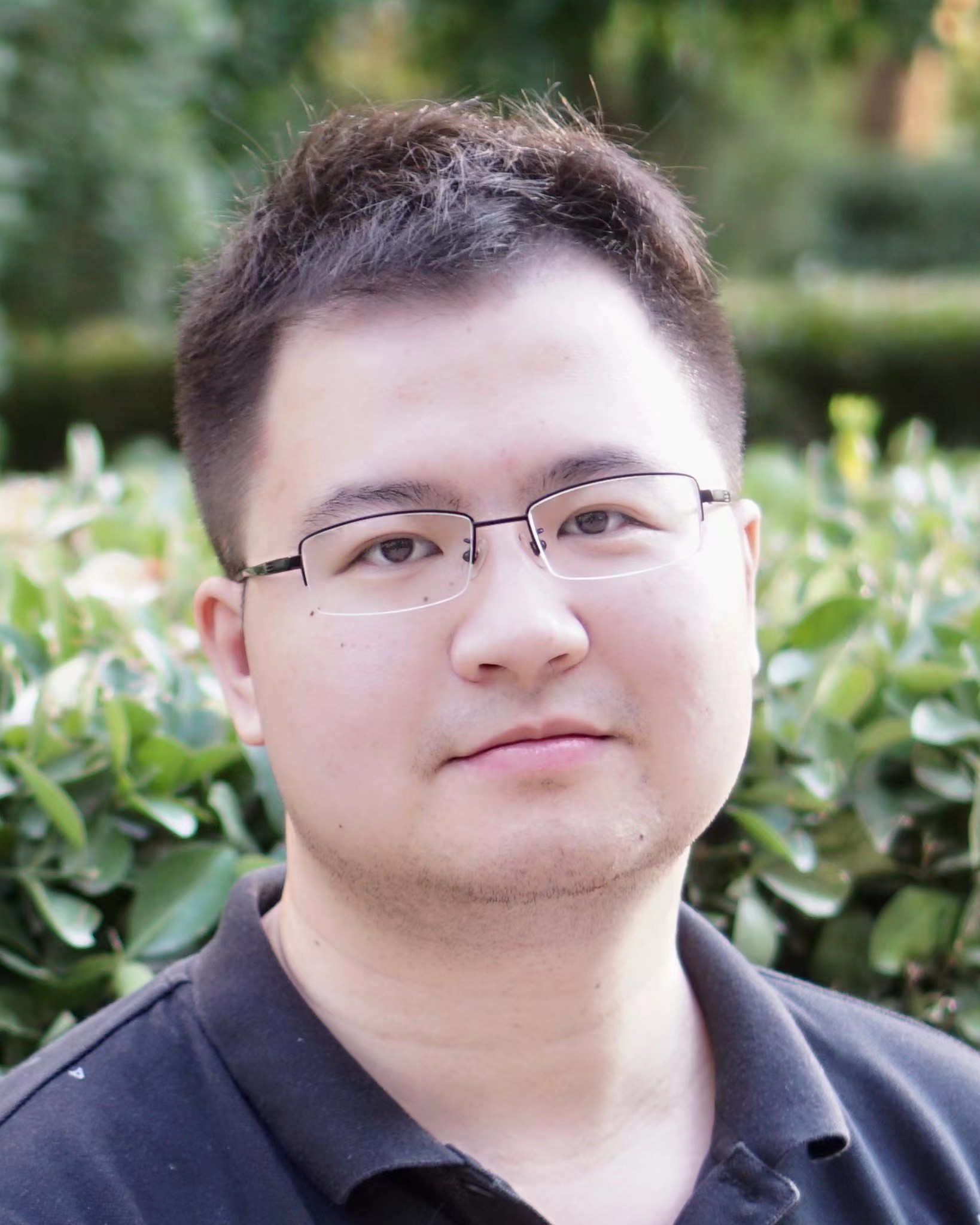}}]{Sheng Yang} received his Ph.D. degree from Tsinghua University in 2019.
		
		He is currently a research engineer leading the localization, calibration and offline-mapping team in the Autonomous Driving Lab of Cainiao, Alibaba Group. His research interests include computer graphics, geometric modeling, and simultaneous localization and mapping for robotics.
	\end{IEEEbiography}
	\begin{IEEEbiography}[{\includegraphics[width=1in,height=1.25in,clip,keepaspectratio]{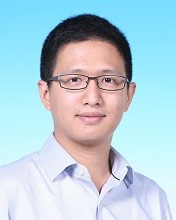}}]{Shaojie Shen} received the B.Eng. degree in electronic engineering from the Hong Kong University of Science and Technology, Hong Kong, in 2009, and the M.S. degree in robotics and the Ph.D. degree in electrical and systems engineering from the University of Pennsylvania, Philadelphia, PA, USA, in 2011 and 2014, respectively.
		
		In September 2014, he joined the Department of Electronic and Computer Engineering, Hong Kong University of Science and Technology, as an Assistant Professor, and was promoted to Associate Professor in July 2020. His research interests include robotics and unmanned aerial vehicles, with a focus on state estimation, sensor fusion, computer vision, localization and mapping, and autonomous navigation in complex environments.
	\end{IEEEbiography}
	
	\begin{IEEEbiography}[{\includegraphics[width=1in,height=1.25in]{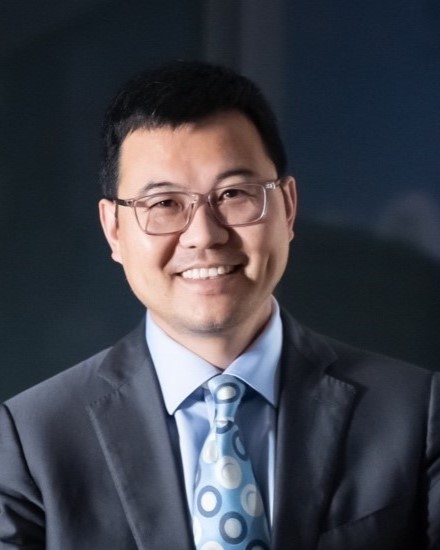}}]{Chao Xu} (Senior Member, IEEE) received the Ph.D. degree in mechanical engineering from Lehigh University in 2010. 
		
		He is currently an Associate Dean and a Professor at the College of Control Science and Engineering, Zhejiang University. He serves as the inaugural Dean	of the ZJU Huzhou Institute. His research expertise is flying robotics and control-theoretic learning.	He has published over 100 articles in international journals, including \textit{Science Robotics} (Cover Paper) and \textit{Nature Machine Intelligence} (Cover Paper).	He will join the organization committee of the IROS-2025, Hangzhou. He was the Managing Editor of \textit{IET Cyber-Systems and Robotics}.
	\end{IEEEbiography}
	
	\begin{IEEEbiography}[{\includegraphics[width=1in,height=1.25in,clip,keepaspectratio]{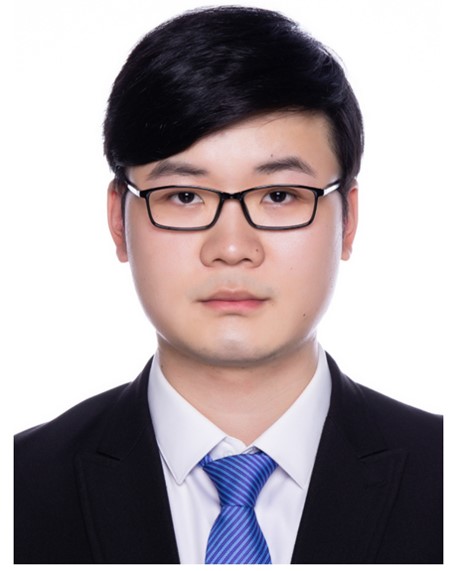}}]{Fei Gao} (Member, IEEE) received the Ph.D. degree in electronic and computer engineering from The Hong Kong University of Science and Technology,
		Hong Kong, in 2019.	
		
		He is currently a tenured Associate Professor at the Department of Control Science and Engineering, Zhejiang University, where he leads the Flying Autonomous Robotics (FAR) Group affiliated with	the Field Autonomous System and Computing (FAST) Laboratory. His research interests include	aerial robots, autonomous navigation, motion planning, optimization, and localization and mapping.
	\end{IEEEbiography}
	\bibliographystyle{./support/IEEEtran}
	
\end{document}

%% file: macros/macro.tex
\usepackage{mathtools}

\newcommand{\Cov}{\boldsymbol{\Sigma}}

\newcommand{\rotvel}{\boldsymbol{\omega}}
\newcommand{\accvel}{\boldsymbol{a}}
\newcommand{\bias}{\boldsymbol{b}}
\newcommand{\bavel}{\bias_{\accvel}}
\newcommand{\bgvel}{\bias_{\rotvel}}
\newcommand{\navel}{\nb_{\accvel}}
\newcommand{\ngvel}{\nb_{\rotvel}}

\newcommand{\bodyframe}{\mathcal{B}}
\newcommand{\graframe}{\mathcal{G}}
\newcommand{\imuframe}{\mathcal{I}}

\DeclarePairedDelimiter\abs{\lvert}{\rvert}%
\DeclarePairedDelimiter\norm{\lVert}{\rVert}%

\makeatletter
\let\oldabs\abs
\def\abs{\@ifstar{\oldabs}{\oldabs*}}
\let\oldnorm\norm
\def\norm{\@ifstar{\oldnorm}{\oldnorm*}}
\makeatother

\usepackage{mathtools}

%% file: paper.bbl
\begin{thebibliography}{10}
\providecommand{\url}[1]{#1}
\csname url@rmstyle\endcsname
\providecommand{\newblock}{\relax}
\providecommand{\bibinfo}[2]{#2}
\providecommand\BIBentrySTDinterwordspacing{\spaceskip=0pt\relax}
\providecommand\BIBentryALTinterwordstretchfactor{4}
\providecommand\BIBentryALTinterwordspacing{\spaceskip=\fontdimen2\font plus
\BIBentryALTinterwordstretchfactor\fontdimen3\font minus
  \fontdimen4\font\relax}
\providecommand\BIBforeignlanguage[2]{{%
\expandafter\ifx\csname l@#1\endcsname\relax
\typeout{** WARNING: IEEEtran.bst: No hyphenation pattern has been}%
\typeout{** loaded for the language `#1'. Using the pattern for}%
\typeout{** the default language instead.}%
\else
\language=\csname l@#1\endcsname
\fi
#2}}

\bibitem{chen2022milestones}
L.~Chen, Y.~Li, C.~Huang, B.~Li, Y.~Xing, D.~Tian, L.~Li, Z.~Hu, X.~Na, Z.~Li,
  \emph{et~al.}, ``Milestones in autonomous driving and intelligent vehicles:
  Survey of surveys,'' \emph{IEEE Transactions on Intelligent Vehicles},
  vol.~8, no.~2, pp. 1046--1056, 2022.

\bibitem{yusefi2023generalizable}
A.~Yusefi, A.~Durdu, F.~Bozkaya, {\c{S}}.~T{\i}{\u{g}}l{\i}o{\u{g}}lu,
  A.~Y{\i}lmaz, and C.~Sungur, ``A generalizable d-vio and its fusion with
  gnss/imu for improved autonomous vehicle localization,'' \emph{IEEE
  Transactions on Intelligent Vehicles}, 2023.

\bibitem{wu2007gps}
B.-F. Wu, T.-T. Lee, H.-H. Chang, J.-J. Jiang, C.-N. Lien, T.-Y. Liao, and
  J.-W. Perng, ``Gps navigation based autonomous driving system design for
  intelligent vehicles,'' in \emph{2007 IEEE International Conference on
  Systems, Man and Cybernetics}.\hskip 1em plus 0.5em minus 0.4em\relax IEEE,
  2007, pp. 3294--3299.

\bibitem{li2020lidar}
Y.~Li and J.~Ibanez-Guzman, ``Lidar for autonomous driving: The principles,
  challenges, and trends for automotive lidar and perception systems,''
  \emph{IEEE Signal Processing Magazine}, vol.~37, no.~4, pp. 50--61, 2020.

\bibitem{loam}
J.~Zhang and S.~Singh, ``{LOAM}: Lidar odometry and mapping in real-time,''
  \emph{Proceedings of Robotics: Science and Systems Conference}, July 2014.

\bibitem{lin2020loam}
J.~Lin and F.~Zhang, ``Loam livox: A fast, robust, high-precision lidar
  odometry and mapping package for lidars of small fov,'' in \emph{2020 IEEE
  International Conference on Robotics and Automation (ICRA)}.\hskip 1em plus
  0.5em minus 0.4em\relax IEEE, 2020, pp. 3126--3131.

\bibitem{geiger2012we}
A.~Geiger, P.~Lenz, and R.~Urtasun, ``Are we ready for autonomous driving? the
  kitti vision benchmark suite,'' in \emph{2012 IEEE conference on computer
  vision and pattern recognition}.\hskip 1em plus 0.5em minus 0.4em\relax IEEE,
  2012, pp. 3354--3361.

\bibitem{davison2007monoslam}
A.~J. Davison, I.~D. Reid, N.~D. Molton, and O.~Stasse, ``Monoslam: Real-time
  single camera slam,'' \emph{IEEE transactions on pattern analysis and machine
  intelligence}, vol.~29, no.~6, pp. 1052--1067, 2007.

\bibitem{klein2007parallel}
G.~Klein and D.~Murray, ``Parallel tracking and mapping for small ar
  workspaces,'' in \emph{2007 6th IEEE and ACM international symposium on mixed
  and augmented reality}.\hskip 1em plus 0.5em minus 0.4em\relax IEEE, 2007,
  pp. 225--234.

\bibitem{mur2015orb}
R.~Mur-Artal, J.~M.~M. Montiel, and J.~D. Tardos, ``{ORB-SLAM}: a versatile and
  accurate monocular slam system,'' \emph{IEEE transactions on robotics},
  vol.~31, no.~5, pp. 1147--1163, 2015.

\bibitem{forster2014svo}
C.~Forster, M.~Pizzoli, and D.~Scaramuzza, ``{SVO}: Fast semi-direct monocular
  visual odometry,'' in \emph{2014 IEEE international conference on robotics
  and automation (ICRA)}.\hskip 1em plus 0.5em minus 0.4em\relax IEEE, 2014,
  pp. 15--22.

\bibitem{yang2019observability}
Y.~Yang and G.~Huang, ``Observability analysis of aided ins with heterogeneous
  features of points, lines, and planes,'' \emph{IEEE Transactions on
  Robotics}, vol.~35, no.~6, pp. 1399--1418, 2019.

\bibitem{wang2024four}
Y.~Wang, W.~Song, Y.~Zhang, F.~Huang, Z.~Tu, R.~Li, S.~Zhang, and Y.~Lou,
  ``Four years of multimodal odometry and mapping on the rail vehicles,''
  \emph{Journal of Field Robotics}, vol.~41, no.~2, pp. 227--257, 2024.

\bibitem{jeong2019complex}
J.~Jeong, Y.~Cho, Y.-S. Shin, H.~Roh, and A.~Kim, ``Complex urban dataset with
  multi-level sensors from highly diverse urban environments,'' \emph{The
  International Journal of Robotics Research}, vol.~38, no.~6, pp. 642--657,
  2019.

\bibitem{wisth2021unified}
D.~Wisth, M.~Camurri, S.~Das, and M.~Fallon, ``Unified multi-modal landmark
  tracking for tightly coupled lidar-visual-inertial odometry,'' \emph{IEEE
  Robotics and Automation Letters}, vol.~6, no.~2, pp. 1004--1011, 2021.

\bibitem{wu2017vins}
K.~J. Wu, C.~X. Guo, G.~Georgiou, and S.~I. Roumeliotis, ``Vins on wheels,'' in
  \emph{2017 IEEE International Conference on Robotics and Automation
  (ICRA)}.\hskip 1em plus 0.5em minus 0.4em\relax IEEE, 2017, pp. 5155--5162.

\bibitem{zhang2019vision}
M.~Zhang, Y.~Chen, and M.~Li, ``Vision-aided localization for ground robots,''
  in \emph{2019 IEEE/RSJ International Conference on Intelligent Robots and
  Systems (IROS)}.\hskip 1em plus 0.5em minus 0.4em\relax IEEE, 2019, pp.
  2455--2461.

\bibitem{kang2019vins}
R.~Kang, L.~Xiong, M.~Xu, J.~Zhao, and P.~Zhang, ``Vins-vehicle: A
  tightly-coupled vehicle dynamics extension to visual-inertial state
  estimator,'' in \emph{2019 IEEE Intelligent Transportation Systems Conference
  (ITSC)}.\hskip 1em plus 0.5em minus 0.4em\relax IEEE, 2019, pp. 3593--3600.

\bibitem{xiao2021lio}
H.~Xiao, Y.~Han, J.~Zhao, J.~Cui, L.~Xiong, and Z.~Yu, ``Lio-vehicle: A
  tightly-coupled vehicle dynamics extension of lidar inertial odometry,''
  \emph{IEEE Robotics and Automation Letters}, vol.~7, no.~1, pp. 446--453,
  2021.

\bibitem{brossard2019rins}
M.~Brossard, A.~Barrau, and S.~Bonnabel, ``Rins-w: Robust inertial navigation
  system on wheels,'' in \emph{2019 IEEE/RSJ International Conference on
  Intelligent Robots and Systems (IROS)}.\hskip 1em plus 0.5em minus
  0.4em\relax IEEE, 2019, pp. 2068--2075.

\bibitem{brossard2020ai}
------, ``Ai-imu dead-reckoning,'' \emph{IEEE Transactions on Intelligent
  Vehicles}, vol.~5, no.~4, pp. 585--595, 2020.

\bibitem{ronin2020Herath}
S.~{Herath}, H.~{Yan}, and Y.~{Furukawa}, ``{RoNIN}: Robust neural inertial
  navigation in the wild: Benchmark, evaluations, new methods,'' \emph{2020
  IEEE Int. Comf. Robot. Autom. (ICRA)}, pp. 3146--3152, 2020.

\bibitem{tlio2020liu}
\BIBentryALTinterwordspacing
W.~Liu, D.~Caruso, E.~Ilg, J.~Dong, A.~Mourikis, K.~Daniilidis, V.~Kumar,
  J.~Engel, A.~Valada, and T.~Asfour, ``{TLIO}: Tight learned inertial
  odometry,'' \emph{IEEE Robotics and Automation Letters}, p. 1–1, 2020.
  [Online]. Available: \url{http://dx.doi.org/10.1109/LRA.2020.3007421}
\BIBentrySTDinterwordspacing

\bibitem{brossard2019learning}
M.~Brossard and S.~Bonnabel, ``Learning wheel odometry and imu errors for
  localization,'' in \emph{2019 IEEE Int. Comf. Robot. Autom. (ICRA)}, 2019,
  pp. 291--297.

\bibitem{zhang2021pose}
M.~Zhang, X.~Zuo, Y.~Chen, Y.~Liu, and M.~Li, ``Pose estimation for ground
  robots: On manifold representation, integration, reparameterization, and
  optimization,'' \emph{IEEE Transactions on Robotics}, vol.~37, no.~4, pp.
  1081--1099, 2021.

\bibitem{9347674}
M.~Ouyang, Z.~Cao, P.~Guan, Z.~Li, C.~Zhou, and J.~Yu, ``Visual-gyroscope-wheel
  odometry with ground plane constraint for indoor robots in dynamic
  environment,'' \emph{IEEE Sensors Letters}, vol.~5, no.~3, pp. 1--4, 2021.

\bibitem{10105944}
J.~Chen, H.~Wang, M.~Hu, and P.~N. Suganthan, ``Versatile lidar-inertial
  odometry with se(2) constraints for ground vehicles,'' \emph{IEEE Robotics
  and Automation Letters}, vol.~8, no.~6, pp. 3486--3493, 2023.

\bibitem{gordon1974b}
W.~J. Gordon and R.~F. Riesenfeld, ``B-spline curves and surfaces,'' in
  \emph{Computer aided geometric design}.\hskip 1em plus 0.5em minus
  0.4em\relax Elsevier, 1974, pp. 95--126.

\bibitem{lee2020visual}
W.~Lee, K.~Eckenhoff, Y.~Yang, P.~Geneva, and G.~Huang, ``Visual-inertial-wheel
  odometry with online calibration,'' in \emph{2020 IEEE/RSJ International
  Conference on Intelligent Robots and Systems (IROS)}.\hskip 1em plus 0.5em
  minus 0.4em\relax IEEE, 2020, pp. 4559--4566.

\bibitem{liang2002visual}
B.~Liang and N.~Pears, ``Visual navigation using planar homographies,'' in
  \emph{Proceedings 2002 IEEE International Conference on Robotics and
  Automation (Cat. No. 02CH37292)}, vol.~1.\hskip 1em plus 0.5em minus
  0.4em\relax IEEE, 2002, pp. 205--210.

\bibitem{scaramuzza20111}
D.~Scaramuzza, ``1-point-ransac structure from motion for vehicle-mounted
  cameras by exploiting non-holonomic constraints,'' \emph{International
  journal of computer vision}, vol.~95, no.~1, pp. 74--85, 2011.

\bibitem{svacha2019inertial}
J.~Svacha, G.~Loianno, and V.~Kumar, ``Inertial yaw-independent velocity and
  attitude estimation for high-speed quadrotor flight,'' \emph{IEEE Robotics
  and Automation Letters}, vol.~4, no.~2, pp. 1109--1116, 2019.

\bibitem{zhang2022dido}
K.~Zhang, C.~Jiang, J.~Li, S.~Yang, T.~Ma, C.~Xu, and F.~Gao, ``Dido: Deep
  inertial quadrotor dynamical odometry,'' \emph{IEEE Robotics and Automation
  Letters}, vol.~7, no.~4, pp. 9083--9090, 2022.

\bibitem{he2016deep}
K.~He, X.~Zhang, S.~Ren, and J.~Sun, ``Deep residual learning for image
  recognition,'' in \emph{CVPR}, 2016.

\bibitem{sola2017quaternion}
J.~Sola, ``Quaternion kinematics for the error-state kalman filter,''
  \emph{arXiv preprint arXiv:1711.02508}, 2017.

\bibitem{carlevaris2016university}
N.~Carlevaris-Bianco, A.~K. Ushani, and R.~M. Eustice, ``University of michigan
  north campus long-term vision and lidar dataset,'' \emph{The International
  Journal of Robotics Research}, vol.~35, no.~9, pp. 1023--1035, 2016.

\bibitem{grupp2017evo}
M.~Grupp, ``evo: Python package for the evaluation of odometry and slam.''
  \url{https://github.com/MichaelGrupp/evo}, 2017.

\end{thebibliography}
